\documentclass[10pt,twocolumn,letterpaper]{article}

\usepackage[pagenumbers]{cvpr} 
\usepackage{pifont}
\usepackage{subcaption}
\usepackage{amssymb}
\usepackage{multirow}
\usepackage[table]{xcolor}
\usepackage{makecell}
\usepackage{tcolorbox}
\usepackage{tabularx}
\definecolor{headgray}{gray}{0.9}

\usepackage{arydshln}
\makeatletter
\newcommand{\thickhline}{%
    \noalign {\ifnum 0=`}\fi \hrule height 0.7pt
    \futurelet \reserved@a \@xhline
}

\definecolor{cvprblue}{rgb}{0.21,0.49,0.74}
\usepackage[pagebackref,breaklinks,colorlinks,allcolors=cvprblue]{hyperref}

\def\confName{CVPR}
\def\confYear{2027}

\title{Speech-based Person Retrieval}

\author{
Yucheng Ji$^{1}$ \quad
Yang Li$^{1}$ \quad
Yueqiao Tang$^{1}$ \quad
Yang Bai$^{2}$\\[1pt]
Xinyu Zhou$^{1}$ \quad
Mang Ye$^{2}$ \quad
Min Cao$^{1}$\\[2pt]
$^{1}$School of Computer Science and Technology,
Soochow University, Suzhou, China\\
$^{2}$School of Computer Science,
Wuhan University, Wuhan, China\\
{\tt\small 20254227067@stu.suda.edu.cn, mcao@suda.edu.cn}\\[2pt]
{\small\hypersetup{urlcolor=magenta}\url{https://github.com/Flame-Chasers/Speech-Based-Person-Retrieval}}
}

\begin{document}
\maketitle
\begin{abstract}
While image-based person re-identification (ReID) and text-based person retrieval (TBPR) have achieved remarkable progress, acquiring target images or well-formed textual queries remains impractical in many real-world deployment scenarios. To bridge this gap, we introduce Speech-based Person Retrieval (\textbf{SPR}), a practical yet underexplored task that retrieves target pedestrians directly from spoken descriptions. 
First, we construct \textbf{Speech-PEDES}, the first large-scale benchmark for SPR. Curated through rigorous quality control, it comprises 37,128 images and 74,282 spoken descriptions across 12,003 identities, recorded by 912 speakers from 13 countries and regions. 
Second, we propose Progressive Speech-Person Alignment (\textbf{Pro-SPA}), designed to handle acoustic variability and modality discrepancies. Specifically, Pro-SPA integrates Identity-aware Speech Adaptation (ISA) to learn speaker-invariant global semantics, Local Speech-Segment Alignment (LSSA) to associate fine-grained appearance cues without explicit word-level segmentation, and Dual-Granularity Score Calibration (DGSC) to refine retrieval rankings. 
Extensive experiments validate the effectiveness of our method. 
Overall, we establish a standardized pipeline across data, modeling, and evaluation to serve as a reproducible baseline for future SPR research. 
The code and the Speech-PEDES dataset are publicly available.
\end{abstract}

\vspace{-4pt}
\section{Introduction}
\label{sec:intro}

Given the paramount importance of ensuring public safety through intelligent video surveillance, there is a substantial need to develop systems capable of accurately retrieving a target person from a large image gallery. 
In this context, image-based person re-identification (ReID)~\cite{Liu2024,Yang_2025_CVPR,ye2025transformer,10882953} employs a person image as the query to perform the retrieval, but such a query image is not always readily available in real-world scenarios.
Text-based person retrieval (TBPR)~\cite{Bai2023RaSa,Cao2024TBPSCLIP,Liu2025DMAdapter,ye2026gtr} addresses this limitation by retrieving person images using textual appearance descriptions.

\begin{figure}[t]
    \centering
    \setlength{\abovecaptionskip}{3pt}
    \setlength{\abovecaptionskip}{3pt}
    \includegraphics[width=\linewidth]{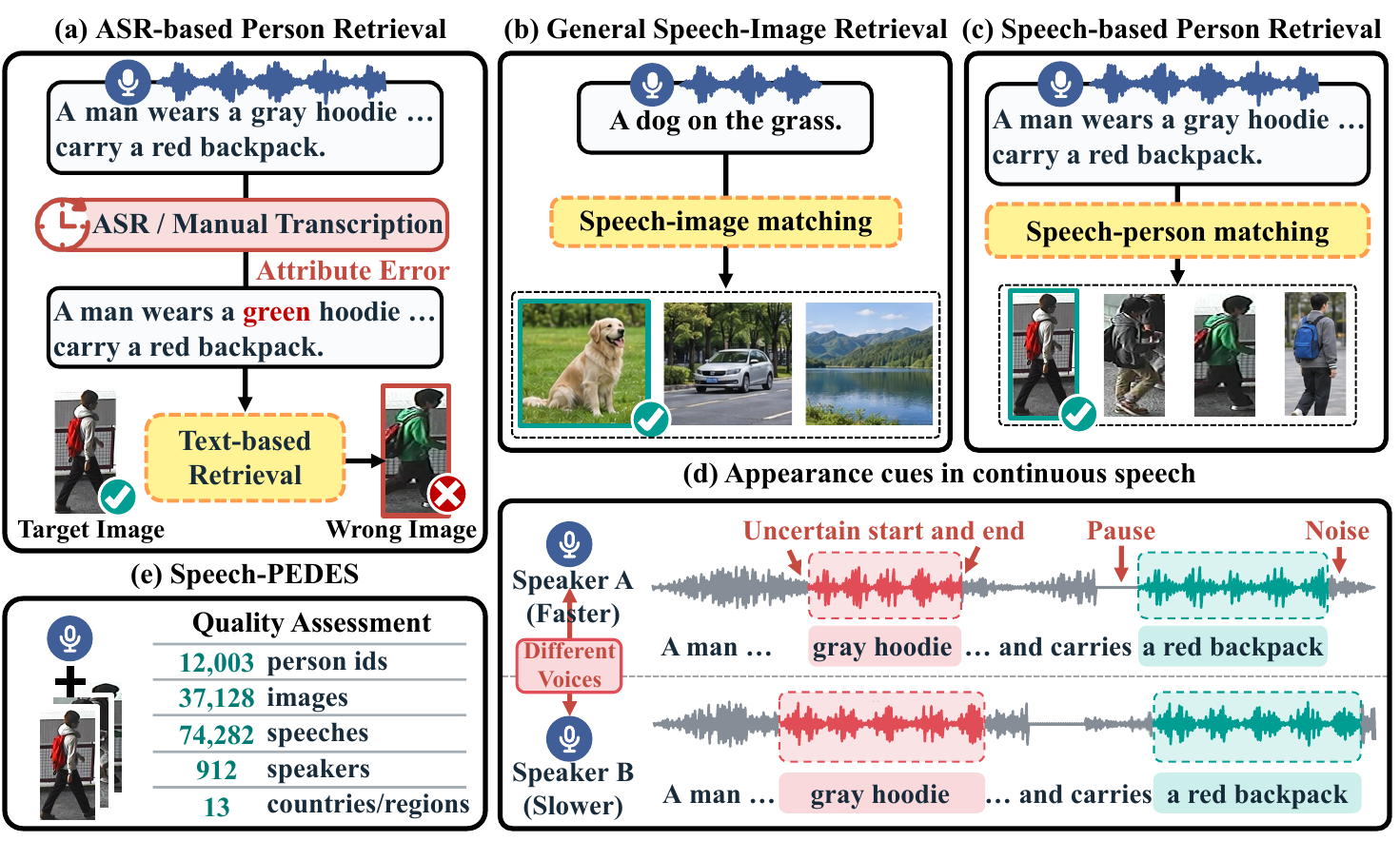}
    \caption{Overview of the proposed SPR paradigm and dataset. (a-c) Comparison of different person retrieval paradigms. (d) Illustration of the inherent challenges in continuous speech. (e) Statistics of the Speech-PEDES dataset.}
  \label{fig:intro}
  \vspace{-18pt}
\end{figure}

In practice, however, well-formed textual descriptions of a target person are not always available, whereas spoken descriptions constitute a more natural and convenient query modality. 
Adapting existing TBPR systems to spoken queries typically requires an intermediate transcription step, either through manual annotation or automatic speech recognition (ASR). 
Manual transcription introduces additional user effort and interaction latency, while ASR outputs may contain recognition errors that distort or omit fine-grained appearance attributes (Fig.~\ref{fig:intro}(a)). 
Consequently, speech-based person retrieval is an important yet underexplored problem setting. 
Although several studies have investigated speech–image retrieval~\cite{Peng2022FaSTVGS,Shih2022SpeechCLIP,Wang2024SpeechCLIPPlus}, they primarily target general natural scenes (\emph{e.g.}, “a dog on the grass”, Fig.~\ref{fig:intro}(b)) and thus focus on coarse-grained semantic alignment. 
These methods do not transfer well to person retrieval, which requires distinguishing visually similar individuals based on subtle, fine-grained appearance cues.

To this end, this work introduces \textbf{Speech-based Person Retrieval (SPR)}, a new task in which spoken appearance descriptions serve directly as queries for retrieving matching person images from a large-scale gallery (Fig.~\ref{fig:intro}(c)). Despite its practical potential, SPR faces two key challenges:
\ding{182} \textbf{Lack of paired speech–image data for person retrieval.}
Existing person retrieval datasets~\cite{Market-1501,MSMT17,Li_2017_CVPR,ICFG,RSTP} provide person images or text–image pairs, but do not include paired speech–image data. Meanwhile, existing speech–image datasets~\cite{flickr8k-audio,Places-audio,spokencoco,speechcoco} are primarily collected for general natural scenes and are not designed for fine-grained person retrieval. This data scarcity poses a significant barrier to the development of SPR.
\ding{183} \textbf{Difficulty of speech semantic extraction and fine-grained cross-modal alignment.}
Unlike textual queries in TBPR, spoken descriptions are entangled with appearance-irrelevant acoustic variations, such as speaker timbre, background noise, and pauses, which hinder the extraction of robust, identity-discriminative appearance representations. Moreover, the key attributes for distinguishing visually similar individuals are often compressed into short, unsegmented speech segments (Fig.~\ref{fig:intro}(d)). 
The boundaries of these segments are obscured by continuous speech, making their precise start and end positions difficult to localize. Their durations also vary with speaking rate. This lack of explicit boundaries and temporal variability makes it challenging for models to accurately capture and localize fine-grained appearance cues, thereby complicating cross-modal alignment between local speech segments and image regions.

In response to the first challenge, we construct \textbf{Speech-PEDES}, the first dataset tailored for SPR. 
It comprises 12,003 pedestrian identities with 37,128 images, and 74,282 spoken descriptions from 912 speakers across 13 countries and regions (Fig.\ref{fig:intro}(e)).
Importantly, the dataset captures real-world speech variations, including diverse speaking rates, natural pauses, and disfluencies. Each recording has been subjected to rigorous quality assessment based on multiple criteria, such as content fidelity and speech naturalness, ensuring the overall reliability of the dataset.
The resulting dataset provides speech-image pairs for training and evaluation, establishing a foundation for SPR research.

In response to the second challenge, we propose \textbf{Progressive Speech-Person Alignment (Pro-SPA)}.
First, Identity-aware Speech Adaptation (ISA) learns a speaker-robust global representation that focuses on pedestrian appearance, effectively extracting identity-discriminative semantics to establish global speech-image alignment.
Building on these representations, Local Speech-Segment Alignment (LSSA) addresses the unclear boundaries and variable durations of spoken attribute cues by establishing local correspondences between speech and image without requiring explicit word-level segmentation.
It further leverages the consistency of bidirectional correspondences to highlight key appearance cues, thereby better distinguishing pedestrians with similar appearances.
Finally, the Dual-Granularity Score Calibration (DGSC) module integrates global and local matching cues to refine the final retrieval ranking.

Overall, the primary contributions of this work can be summarized as follows:
\begin{itemize}

    \item[\ding{182}] We introduce Speech-based Person Retrieval (SPR), a practical yet underexplored task that fills a critical gap in the person retrieval spectrum, extending beyond the classical ReID and TBPR settings.
    
    \item[\ding{183}] We construct Speech-PEDES, the first benchmark dataset tailored for the SPR task, providing a reliable evaluation cornerstone for future research.
    
    \item[\ding{184}] We propose Pro-SPA, which extracts speaker-robust appearance semantics and achieves precise local alignment directly from unsegmented speech.

    \item[\ding{185}] Comprehensive experiments on Speech-PEDES demonstrate the capabilities of Pro-SPA, establishing a simple, effective, and reproducible baseline for the SPR task.
\end{itemize}

\begin{table*}[!t]
\centering
\footnotesize
\setlength{\abovecaptionskip}{3pt}
\setlength{\belowcaptionskip}{3pt}
\caption{Comparison of Speech-PEDES with speech-image retrieval datasets, together with its speech quality. For textual statistics, \#Words, \#Adj., and \#NCs. denote the average number of words, adjectives, and noun chunks per description, respectively. Regarding speech quality evaluated via DNSMOS P.835, SIG, BAK, and OVRL represent the signal, background noise, and overall quality scores.}

\setlength{\tabcolsep}{3pt}
\renewcommand{\arraystretch}{1.2}
\resizebox{\textwidth}{!}{%
\begin{tabular}{c||cc|cccc|cccc|c||ccc|c}
\hline\thickhline
\rowcolor{headgray}
& & & & & & & & & & &
& \multicolumn{3}{c|}{DNSMOS P.835 $\uparrow$}
& \\
\rowcolor{headgray}
\multirow{-2}{*}{Dataset}
& \multirow{-2}{*}{Source}
& \multirow{-2}{*}{Type}
& \multirow{-2}{*}{Speech}
& \multirow{-2}{*}{Image}
& \multirow{-2}{*}{IDs}
& \multirow{-2}{*}{Speaker}
& \multirow{-2}{*}{Duration}
& \multirow{-2}{*}{\#Words}
& \multirow{-2}{*}{\#Adj.}
& \multirow{-2}{*}{\#NCs.}
& \multirow{-2}{*}{Acc}
& SIG & BAK & OVRL
& \multirow{-2}{*}{\makecell{OVRL\\$\geq 3.0$}} \\
\hline\hline
\rowcolor{white}
Flickr8k Audio~\cite{flickr8k-audio}
& Flickr8k~\cite{Flickr8k}
& Scripted
& 40,000 & 8,000 & -- & 183
& 4.33 & 10.81 & 1.02 & 3.18 & 90.40\%
& 3.42 & 3.65 & 2.97 & 52.14\% \\

Places Audio~\cite{Places-audio}
& Places 205~\cite{Places205}
& Unscripted
& 403,385 & 402,826 & -- & 2,683
& 8.37 & 19.30 & 1.71 & 5.76 & 67.62\%
& 3.35 & 3.65 & 2.90 & 45.76\% \\

SpokenCOCO~\cite{spokencoco}
& MS COCO~\cite{MS-COCO}
& Scripted
& 605,495 & 123,287 & -- & 2,353
& 4.12 & 10.45 & 0.76 & 3.27 & 86.10\%
& 3.39 & 3.67 & 2.94 & 48.99\% \\

SpeechCOCO~\cite{speechcoco}
& MS COCO~\cite{MS-COCO}
& Generation
& 616,767 & 123,287 & -- & 8
& 3.53 & 10.46 & 0.76 & 3.27 & 83.40\%
& 3.52 & 4.07 & 3.25 & 91.77\% \\

\hline
\rowcolor[HTML]{D7F6FF} 
\textbf{Speech-PEDES (Ours)}
& CUHK-PEDES~\cite{Li_2017_CVPR}
& \makecell{Scripted\\+ Generation}
& 74,282 & 37,128 & 12,003 & 912
& 8.56 & 23.58 & 4.67 & 6.89 & \textbf{93.97\%}
& \textbf{3.54} & \textbf{4.11} & \textbf{3.29} & \textbf{95.09\%} \\

\hline\thickhline
\end{tabular}%
}
\label{tab:dataset_compare}
\vspace{-14pt}
\end{table*}

\vspace{-3pt}
\section{Related Work}
\label{sec:formatting}

\subsection{Text-based Person Retrieval}
Text-based person retrieval (TBPR) aims to retrieve images of a target person from an image gallery using natural language queries. 
Early methods primarily focused on aligning global image and text representations, and gradually evolved to multi-granularity semantic alignment and local feature modeling~\cite{Li_2017_CVPR,Niu2020MIA,Wang2020ViTAA,Chen2022TIPCB,Shu2022IVT,Zhang2018CMPM,Jing2020PMA,chat}.
With advances in vision–language pretraining, a growing body of research leverages pretrained models to improve cross-modal retrieval~\cite{Jiang2023IRRA,Bai2023RaSa,Zuo2024PLIP,Shao2023UniPT,Yang2023APTM,Yan2023CFine,bai2026interactive,ye2026gtr}. 
Additionally, subsequent studies have explored efficient adaptation methods specifically tailored for person retrieval~\cite{Cao2024TBPSCLIP,Liu2025DMAdapter,Sun2025FilteringWoRA}. 
Recent studies have further investigated noise in image–text matching~\cite{Qin2024RDE,Zheng2025GADMS,Yuan2026GSCA,You2026KPDM} and fine-grained differences among visually similar pedestrians~\cite{Park2024PLOT,Zou2026CECA,Li2026QCAlign,Wang2024PersonSAM,Liang2026DFSM,zhou2026vision,zhou2026text,cao2025multilingual}.
Despite these advances, well-formed textual descriptions are not always readily available in real-world scenarios, whereas spoken descriptions constitute a more natural and convenient query modality. 

\vspace{-3pt}
\subsection{Speech-Image Retrieval}
Speech-image retrieval directly retrieve images from spoken descriptions. 
To establish a shared cross-modal semantic space, foundational studies have explored dual--encoder architectures and pretrained models, such as MILAN~\cite{Sanabria2021MILAN}, FaST-VGS~\cite{Peng2022FaSTVGS}, and the SpeechCLIP series~\cite{Shih2022SpeechCLIP,Wang2024SpeechCLIPPlus}. More recently, efforts have shifted toward refining cross-modal representations through advanced alignment strategies and novel training objectives, including coarse--to--fine matching~\cite{Zhou2024Coarse} and cross-modal denoising~\cite{Zhou2024CMD}.
Different from existing speech-image retrieval methods that emphasize general natural scenes and coarse-grained semantic alignment, this work focuses on the specific field of SPR. It requires the model to retrieve the target pedestrian from an image gallery by capturing fine-grained appearance cues. While an early demo~\cite{chen2020cross} explored this, it relies on intermediate transcriptions rather than direct speech-image retrieval.

\vspace{-3pt}
\subsection{Omni-Modal Models}


Recent years have witnessed remarkable progress in omni-modal models, which process heterogeneous inputs including text, images, speech, and video within a unified framework. Foundational works, such as VITA\cite{Fu2024VITA}, Baichuan-Omni-1.5\cite{Li2025BaichuanOmni}, and the Qwen-Omni series~\cite{Xu2025Qwen25Omni,Xu2025Qwen3Omni,team2026qwen3}, have successfully integrated visual and spoken perception into end-to-end language architectures. Concurrent efforts like OmniVinci~\cite{Ye2026OmniVinci} further scale these capabilities through joint semantic alignment and temporal modeling. Beyond general multimodal understanding, recent studies~\cite{Wang2025OmniBind,Ovis2026Embedding,Tang2026WAVE,Huynh2026OmniRet} have begun to explore unified representations for cross-modal retrieval based on omni-modal models.



\vspace{-5pt}
\section{Speech-PEDES}
\label{sec:benchmark}

To address the lack of paired speech-image data for SPR, we construct \textbf{Speech-PEDES}, a large-scale dataset tailored for SPR tasks in real-world scenarios. Rigorous quality assessments (Sec.~\ref{sec:quality} and Appendix.~\ref{app:data_assessment}) indicate that Speech-PEDES provides reliable speech-image pairs.

\begin{figure}[t]
    \centering
    \setlength{\abovecaptionskip}{3pt}
    \setlength{\belowcaptionskip}{3pt}
    \includegraphics[width=\linewidth]{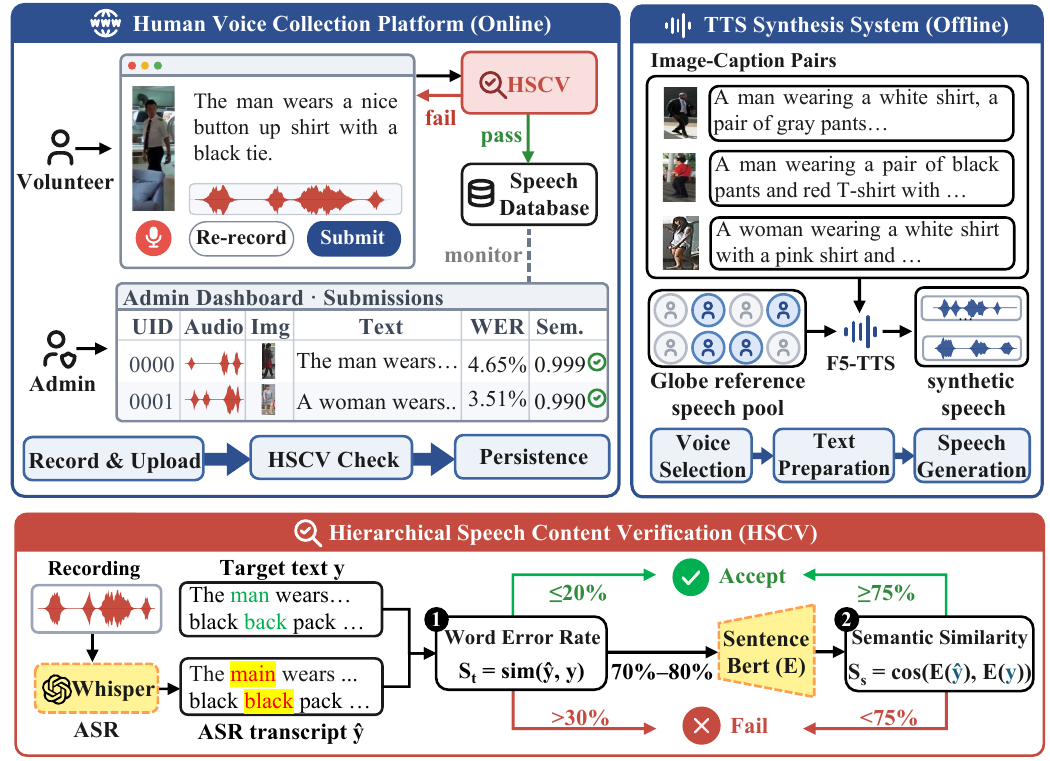}
    \caption{Overview of the Speech-PEDES construction pipeline.}
    \label{fig:data_construction}
    \vspace{-14pt}
\end{figure}

\vspace{-4pt}
\subsection{Data Construction}

We develop a dedicated web-based recording platform (details are in Appendix~\ref{app:recording_platform}) and recruited 623 volunteers across 13 countries and regions to record spoken descriptions based on pedestrian images and reference captions from CUHK-PEDES~\cite{Li_2017_CVPR}.
As shown in Fig.~\ref{fig:data_construction}, to ensure recording quality during collection, we propose a \textbf{Hierarchical Speech Content Verification (HSCV)} mechanism. This mechanism combines word error rate (WER) and semantic similarity to verify consistency between spoken recordings and their corresponding reference captions.
Specifically, we employ Whisper~\cite{whisper} to transcribe each recording into text and calculate the WER against the reference caption. WER effectively identifies and filters out recordings suffering from severe acoustic noise, mispronunciations, or significant script deviations. Recordings with $\text{WER} \leq 20\%$ are directly retained, whereas those with $\text{WER} > 30\%$ are rejected for re-recording. For the remaining recordings ($20\% < \text{WER} \leq 30\%$), we utilize Sentence-BERT~\cite{sentence-bert} to extract semantic representations and compute the cosine similarity with the reference caption. This semantic-level verification effectively filters out recordings where the lexical errors lead to the loss or alteration of critical appearance attributes (e.g., clothe colors), retaining only those with a similarity score $\geq 0.75$. 

To further mitigate data scarcity and enhance speaker diversity, we complement real-world recordings with high-fidelity synthesized speech. Specifically, we sample 289 high-quality reference speakers across 12 regions from GLOBE~\cite{globe} to serve as timbre sources and employ F5-TTS model~\cite{f5-tts} to generate speech from reference captions. 
To ensure these synthetic samples faithfully reflect real-world complexities, we go beyond standard text-to-speech by incorporating dynamic prosodic features, including variable speaking rates and spontaneous pauses (see Appendix~\ref{app:speech_synthesis} for details).
This augmentation strategy not only alleviates the data bottleneck but also enhances the model's robustness against the variability found in natural human speech.

Through this pipeline, we obtain 37,128 images of 12,003 pedestrian identities and 74,282 spoken queries from 912 speakers across 13 regions (as shown in Tab.~\ref{tab:dataset_compare}). Sample visualization is presented in Appendix~\ref{app:dataset_examples}.

\vspace{-3pt}
\subsection{Quality Assessment}
\label{sec:quality}
\begin{figure}[t]
    \centering
    \setlength{\abovecaptionskip}{3pt}
    \setlength{\belowcaptionskip}{3pt}
    \includegraphics[width=0.95\linewidth]{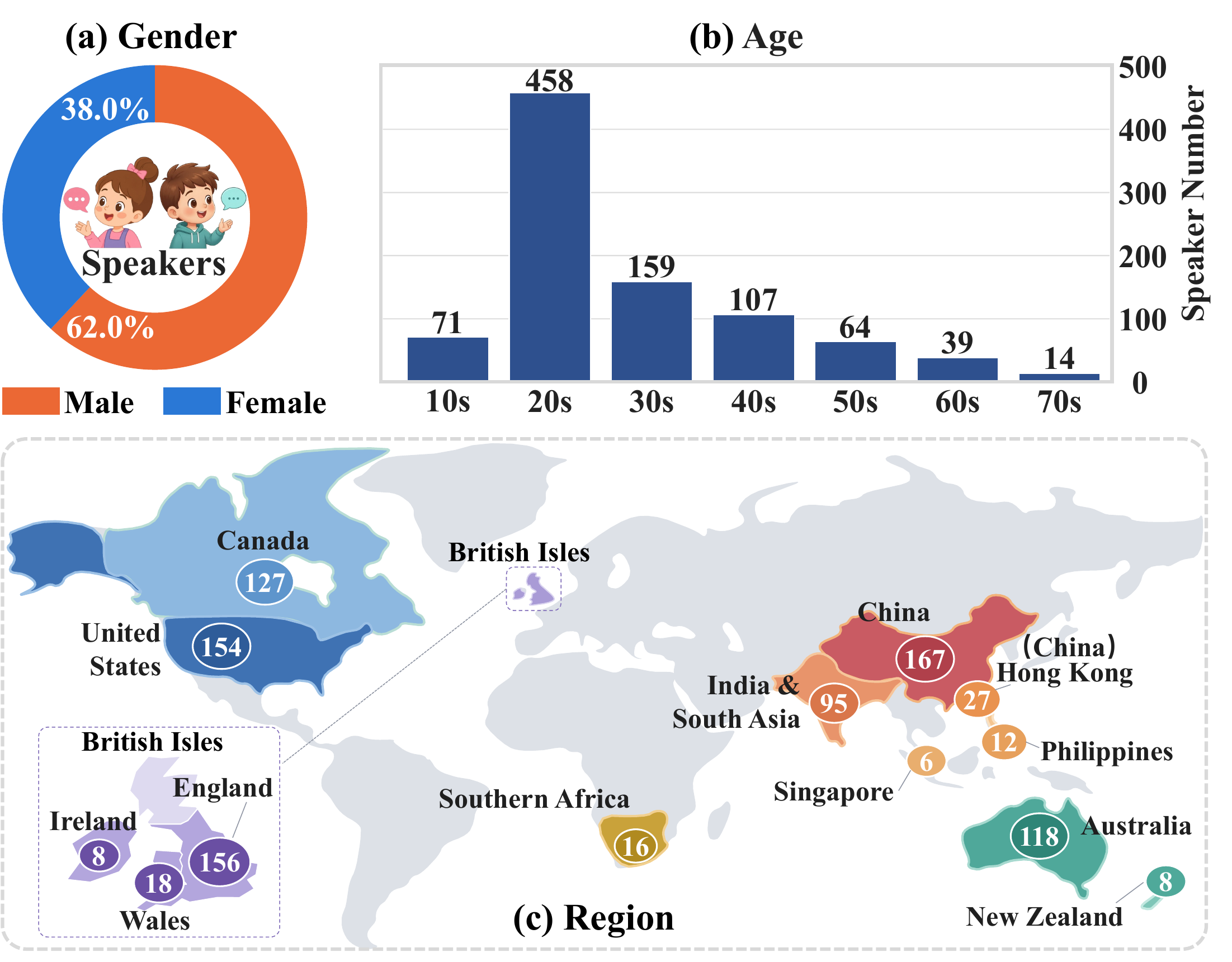}
    \caption{Demographic diversity of speakers in Speech-PEDES, categorized by region, gender, and age.}
    \label{fig:speaker_diversity}
    \vspace{-14pt}
\end{figure}

We conduct a comprehensive evaluation of Speech-PEDES along three dimensions: description richness, speech quality, and speaker diversity. Quantitative and qualitative analyses consistently demonstrate that Speech-PEDES provides reliable speech-image pairs.
\ding{182} For description richness, we measure the average speech duration, word count, adjective count, and noun chunk count. As shown in Tab.~\ref{tab:dataset_compare}, Speech-PEDES has an average speech duration of 8.56 seconds, with each sample containing an average of 23.58 words, 4.67 adjectives, and 6.89 noun chunks. 
Compared with general speech-image retrieval datasets for natural scenes, Speech-PEDES shows longer durations and a markedly higher density of adjectives and noun chunks, providing abundant fine-grained appearance cues for pedestrian identity discrimination.
\ding{183} For speech quality, we use DNSMOS P.835~\cite{dnsmos} to evaluate speech signal quality (SIG), background noise quality (BAK), and overall quality (OVRL).
The results in Tab.\ref{tab:dataset_compare} indicate that Speech-PEDES exhibits highest speech signal fidelity, minimal background noise interference, and superior overall perceptual quality.
Furthermore, we compute speaker similarity (SIM-o) to assess timbral consistency between the synthesized speech and reference audio. The synthesized speech achieves a SIM-o score of 0.87, demonstrating high timbral fidelity.
This result validates the strong voice-cloning capability of F5-TTS~\cite{f5-tts}, showing that modern TTS models are highly capable of synthesizing realistic pedestrian spoken descriptions, as these descriptions of pedestrian appearance contain a narrow semantic scope and limited demands on expressive prosody. This confirms that generative approaches are highly reliable for constructing high-quality speech-image datasets.
\ding{184} Regarding speaker diversity, Speech-PEDES encompasses 912 speakers, with male and female speakers accounting for 62\% and 38\%, respectively (Fig.~\ref{fig:speaker_diversity} (a)), and ages spanning from the 10s to the 70s (Fig.~\ref{fig:speaker_diversity} (b)).
In addition, the dataset covers 13 representative regional accents across Asia, Europe, North America, Africa, and Oceania, capturing the natural pronunciation characteristics of both native and non-native English speakers (Fig.~\ref{fig:speaker_diversity} (c)).

A web demo with playable audio samples is available through an \href{https://anonymous.4open.science/r/SPR-0365/}{anonymous repository}. Additional analyses, covering utterance duration and speaking rate distributions, attribute recognition errors, and comparisons across TTS models, are provided in the Appendix~\ref{app:data_assessment}.


\begin{figure*}[t]
    \centering
    \setlength{\abovecaptionskip}{2pt}
    \setlength{\belowcaptionskip}{2pt}
    \includegraphics[width=0.92\linewidth]{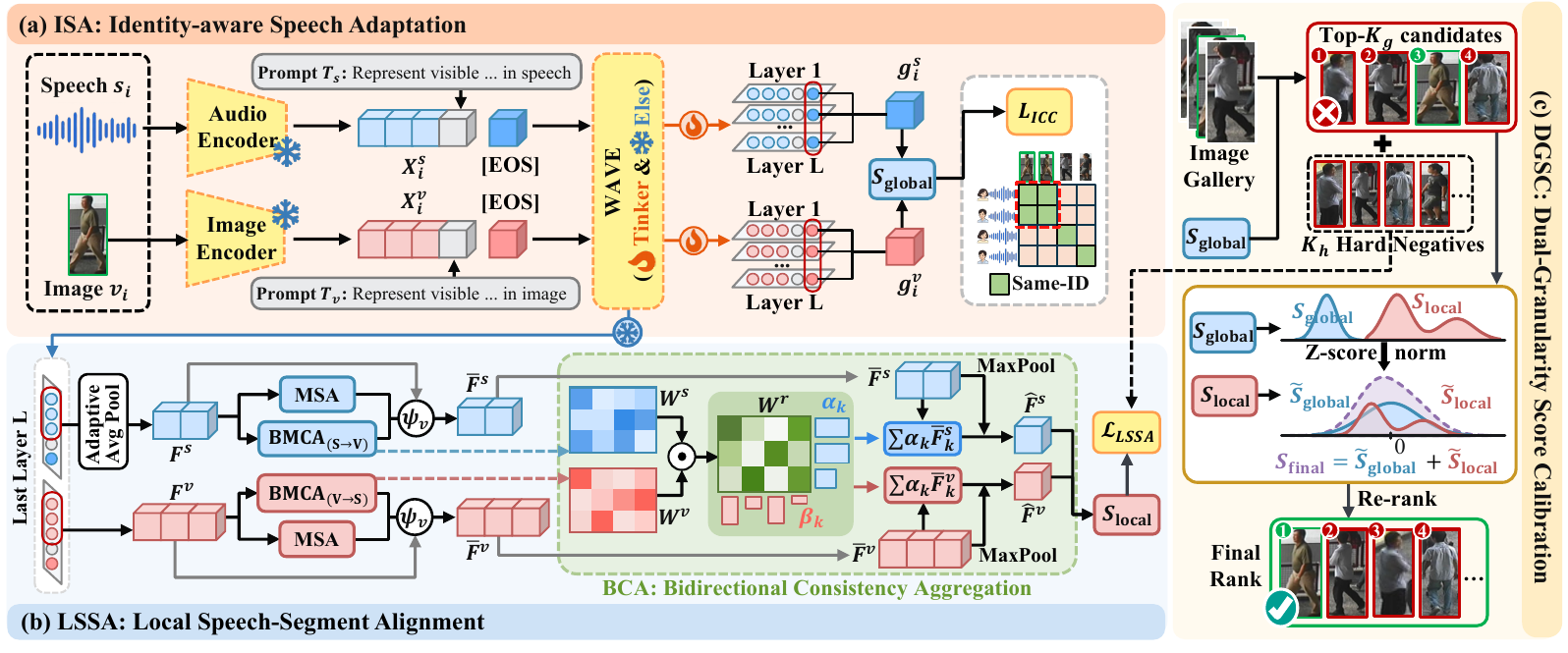}
    \caption{Overview of Pro-SPA. Identity-aware Speech Adaptation (ISA) uses visual identity supervision to learn identity-discriminative appearance semantics from speech. Local Speech-Segment Alignment (LSSA) captures sparse appearance cues by aligning speech segments with image regions without explicit word-level segmentation. During inference, Dual-Granularity Score Calibration (DGSC) selects the top-\(K_g\) candidates using global matching scores and reranks them by summing Z-score-normalized global and local scores.}
    \label{fig:method}
    \vspace{-16pt}
\end{figure*}

\vspace{-4pt}
\section{Method}
\label{sec:method}
\vspace{-2pt}

To address the challenges of speech semantic extraction and fine-grained cross-modal alignment in SPR, we propose the \textbf{Progressive Speech-Person Alignment (Pro-SPA)}. It comprises three core modules.

\vspace{-3pt}
\subsection{ISA: Identity-aware Speech Adaptation}
\label{sec:isa}
Unlike textual queries in TBPR, spoken descriptions naturally carry appearance-irrelevant factors (\emph{e.g.}, speaker timbre, background noise, and pauses). Therefore, an essential challenge in SPR is to decouple the underlying appearance semantics from these acoustic variations.
We therefore propose \textbf{Identity-aware Speech Adaptation (ISA)}, which uses visual identity supervision to align same-identity speech-image pairs and learn identity-discriminative appearance semantics across speaker variation.

As depicted in Fig.~\ref{fig:method}, given a speech query $s_i$, we first encode it into a sequence of speech tokens using the speech encoder of WAVE~\cite{Tang2026WAVE}. 
We then append a person-centric system message $T_s$ (see Appendix~\ref{app:prompts} for details) to these speech tokens to activate the instruction-following capability of WAVE, followed by an end-of-sequence ($\texttt{[EOS]}$) token, forming the final input sequence $X^s_i$. 
For an image $v_j$, the visual input sequence $X_j^v$ is constructed analogously by concatenating WAVE's visual tokens, the person-centric system message $T_v$, and an $\texttt{[EOS]}$ token. Both sequences are then processed by WAVE's shared Thinker. 
Finally, we average the hidden states of the $\texttt{[EOS]}$ token across all $L$ layers of the Thinker and apply $\ell_2$-normalization ($\ell_2(\cdot)$) to obtain the global representations $g_i^s$ and $g_j^v$:
\vspace{-2pt}
\begin{equation}
    g_i^s = \ell_2\left(
    \frac{1}{L}\sum_{\ell=1}^{L}h_{i,\ell}^s
    \right), \quad
    g_j^v = \ell_2\left(
    \frac{1}{L}\sum_{\ell=1}^{L}h_{j,\ell}^v
    \right),
\label{eq:global_embeddings}
\end{equation}
\par
\vspace{-2pt}
\noindent where $h_{i,\ell}^s$ and $h_{j,\ell}^v$ denote the speech and image $\texttt{[EOS]}$ hidden states at layer $\ell$, respectively.
Their cosine similarity defines the global matching score:
\vspace{-3pt}
\begin{equation}
    S_{\text{global}}(s_i,v_j)
    =
    \frac{(g_i^s)^\top g_j^v}
    {\|g_i^s\|_2\|g_j^v\|_2}.
    \label{eq:global_similarity}
\end{equation}
\par
\vspace{-2pt}
\noindent Given this global matching score, conventional cross-modal retrieval methods typically employ single-positive contrastive objectives~\cite{rusak2025infonce}, which align a query with its paired sample while treating all other samples as negatives.
However, such a strict formulation treats other valid samples of the same identity as false negatives, mistakenly pushing them away in the feature space. This may force the model to overfit to specific speech-image pairs rather than learning speaker-invariant visual semantics (see Appendix~\ref{app:robutness_to_vary} and \ref{app:visualization_info_vs_ours} for details).
To overcome this, we formulate the \textbf{Identity-aware Cross-modal Contrastive (ICC)} objective, inspired by supervised contrastive learning~\cite{khosla2021supcon}. ICC treats all speech-image pairs sharing the same identity within a mini-batch as positive matches. By doing so, it forces the model to look beyond appearance-irrelevant factors and extract identity-discriminative appearance semantics across speech queries, thereby achieving a speaker-invariant cross-modal alignment.
Specifically, when constructing a mini-batch of size $B$, we ensure that each speech query is accompanied by at least one additional sample of the same identity, with the remaining samples randomly selected from the dataset.
Let $P_i$ and $P_j$ denote the sets of images and speech samples sharing the same identity, respectively. For brevity, we denote $S_{i,j} = S_{\text{global}}(s_i, v_j)$. The losses are then defined as follows:
\vspace{-1pt} 
\begin{equation}
    \mathcal{L}_{\text{ICC}}^{s \to v} = - \frac{1}{B} \sum_{i=1}^{B} \frac{1}{|P_i|} \sum_{j \in P_i} \log \frac{\exp(S_{i,j} / \tau)}{\sum_{k=1}^{B} \exp(S_{i,k} / \tau)},
    \label{eq:icc_s2v}
\end{equation}
\vspace{-1pt}
\begin{equation}
    \mathcal{L}_{\text{ICC}}^{v \to s} = - \frac{1}{B} \sum_{j=1}^{B} \frac{1}{|P_j|} \sum_{i \in P_j} \log \frac{\exp(S_{i,j} / \tau)}{\sum_{k=1}^{B} \exp(S_{k,j} / \tau)},
    \label{eq:icc_v2s}
\end{equation}
\vspace{-1pt}
\begin{equation}
    \mathcal{L}_{\text{ICC}} = \frac{1}{2} \left( \mathcal{L}_{\text{ICC}}^{s \to v} + \mathcal{L}_{\text{ICC}}^{v \to s} \right),
    \label{eq:icc_loss}
\end{equation}
\par
\vspace{-1pt}
\noindent where $\tau$ is the temperature parameter. 
By aligning same identity speech–image pairs, ICC encourages the model to capture visually grounded appearance semantics shared across diverse spoken descriptions.

\subsection{LSSA: Local Speech-Segment Alignment}
\label{sec:lssa}
The appearance cues that distinguish individuals often occupy only brief portions of continuous speech. Their boundaries are implicit and their durations vary with speaking rate, making these sparse cues difficult to capture within long sequences. To address this, we propose the \textbf{Local Speech-Segment Alignment (LSSA)} mechanism to extract and align these cues without explicit word-level segmentation (see Appendix~\ref{app:local_alignment} for details).

After ISA training, we freeze the encoders, shared Thinker, and global projection heads. 
To compress the length of the speech features and form local matching units without relying on explicit word-level segmentation, we extract the speech-token hidden states from the final layer of the frozen Thinker and apply Adaptive Average Pooling along the temporal dimension, yielding $m$ speech-segment features $F_{\text{loc}}^s$.
For the visual branch, we retain the $n$ final-layer visual-token hidden states as image-region features $F_{\text{loc}}^v$. 
Separate linear projections $\Phi_s$ and $\Phi_v$ then map both modalities to a common feature dimension $d$:
\vspace{-1pt}
\begin{equation}
F^s = \Phi_s(F_{\text{loc}}^s) \in \mathbb{R}^{m \times d}, 
\quad
F^v = \Phi_v(F_{\text{loc}}^v) \in \mathbb{R}^{n \times d}.
\label{eq:local_projection}
\end{equation}
%
\par
\vspace{-1pt}
\noindent Although Adaptive Average Pooling provides a coarse representation of speech, discriminative appearance cues may remain distributed across the resulting speech and visual features.
To selectively capture and integrate these cues, we apply multi-head self-attention (MSA) and bidirectional multi-head cross-attention (BMCA) in parallel to each $(F^s, F^v)$ pair.
Here, MSA enables each local unit to selectively aggregate information from other positions across the sequence. 
BMCA establishes bidirectional interactions between speech segments and image regions, using visual evidence to guide attention toward speech content relevant to pedestrian appearance.
For each modality, we sum the attention outputs, apply a modality-specific projection $\psi$, and add the result to the input features through a residual connection to obtain the updated local feature sequences $\bar{F}^s$ and $\bar{F}^v$:
\vspace{-1pt}
\begin{equation}
\begin{aligned}
    \bar{F}^s &= F^s + \psi_s\big(\text{MSA}(F^s) + \text{BMCA}(F^s, F^v)\big), \\
    \bar{F}^v &= F^v + \psi_v\big(\text{MSA}(F^v) + \text{BMCA}(F^v, F^s)\big).
\end{aligned}
\label{eq:local_refinement}
\end{equation}
\par
\vspace{-1pt}
\noindent These interactions effectively refine the local features. However, not all units in $\bar{F}^s$ and $\bar{F}^v$ are equally informative for identity discrimination. To adaptively emphasize the most informative cues during feature aggregation, we introduce \textbf{Bidirectional Consistency Aggregation (BCA)}, which assigns greater weight to units with stronger reciprocal cross-modal support.
Specifically, we average the speech-to-image and image-to-speech attention maps from BMCA across heads, yielding $W^s \in \mathbb{R}^{m \times n}$ and $W^v \in \mathbb{R}^{n \times m}$, respectively.
Their Hadamard product yields the bidirectional consistency matrix $W^r \in \mathbb{R}^{m \times n}$:
\vspace{-1pt}
\begin{equation}
W^r = W^s \odot (W^v)^\top.
\label{eq:bidirectional_consistency}
\end{equation}
\par
\vspace{-1pt}
\noindent Each $W^r_{(k,l)}$ reflects the mutual attention between $k$-th speech segment and $l$-th image region.
By accumulating these scores along respective dimension and normalizing them globally, we obtain aggregation weights $\alpha_k$ and $\beta_l$:
\vspace{-1pt}
\begin{equation}
    \alpha_k = \frac{1}{Z} \sum_{l=1}^n W_{(k,l)}^r, \quad
    \beta_l = \frac{1}{Z} \sum_{k=1}^m W_{(k,l)}^r,
    \label{eq:aggregation_weights}
\end{equation}
\par
\vspace{-1pt}
\noindent where $Z = \sum_{p=1}^m \sum_{q=1}^n W_{p,q}^r$ serves as the global normalization factor.
We then use $\alpha_k$ and $\beta_l$ to aggregate the updated speech and image features by weighted summation, respectively. 
We also apply Max Pooling to each feature sequence to retain strong local responses.
The weighted and max-pooled representations are then added and passed through layer normalization ($\text{LN}$) to obtain the local matching features $\hat{F}^s$ and $\hat{F}^v$:
\vspace{-1pt}
\begin{equation}
\begin{aligned}
    \hat{F}^s &= \text{LN}\Big(\text{MaxPool}(\bar{F}^s) + \textstyle\sum_{k=1}^m \alpha_k \bar{F}^s_k\Big), \\
    \hat{F}^v &= \text{LN}\Big(\text{MaxPool}(\bar{F}^v) + \textstyle\sum_{l=1}^n \beta_l \bar{F}^v_l\Big).
\end{aligned}
\label{eq:local_aggregation}
\end{equation}
For a speech-image pair $(s_i,v_j)$, we use the cosine similarity of these features to compute the local matching score:
\begin{equation}
    S_{\text{local}}(s_i, v_j) = \frac{(\hat{F}^s)^\top \hat{F}^v}{\Vert{}\hat{F}^s\Vert{}_2 \Vert{}\hat{F}^v\Vert{}_2}.
    \label{eq:local_similarity}
\end{equation}
\par
\vspace{-1pt}
\noindent To further enhance fine-grained discrimination, we train LSSA with hard negatives mined from the training image gallery using the global matching scores $S_{\text{global}}$ provided by the frozen ISA.
For each training query $s_i$, we exclude images sharing its pedestrian identity and select the $K_h$ highest-scoring remaining images to form the hard-negative set $\mathcal{N}_i$.
The positive $v_i^+$ is the image originally paired with $s_i$.
We optimize the following contrastive objective over a mini-batch of $B$ speech queries, encouraging the positive image to receive a higher score than these hard negatives:
\vspace{-1pt}
\begin{equation}
    \mathcal{L}_{\text{LSSA}}
    =
    -\frac{1}{B}\sum_{i=1}^{B}
    \log
    \frac{\exp(S_{\text{local}}(s_i,v_i^+)/\tau_l)}
         {\sum_{v\in\tilde{N}_i}
          \exp(S_{\text{local}}(s_i,v)/\tau_l)},
  \label{eq:lssa_loss}
\end{equation}
\par
\vspace{-1pt}
\noindent where $\tilde{N}_i=\{v_i^+\}\cup\mathcal{N}_i$ and $\tau_l$ is the temperature parameter.

\subsection{DGSC: Dual-Granularity Score Calibration}
\label{sec:dgsc}

During inference, conventional embedding-based retrieval directly ranks candidates by the cosine similarity between speech and image embeddings. However, discriminative appearance cues in spoken descriptions are often sparsely distributed across brief speech segments. These cues may contribute only weakly to direct embedding-level similarity, resulting in limited score separation among visually similar candidates.
We therefore introduce \textbf{Dual-Granularity Score Calibration (DGSC)}, which combines fast global screening with local refinement. DGSC first uses the global matching score $S_{\text{global}}$ from ISA to rapidly narrow the gallery to the top-$K_g$ candidates. It then computes local matching scores $S_{\text{local}}$ using LSSA and combines global and local scores to rerank these candidates, restricting the additional cost of pairwise local interactions to the candidate set.
Specifically, for each speech query $s_i$, let $\mathcal{C}_i$ denote the set of top-$K_g$ candidate images. To align the scales of the global and local scores, we apply query-specific Z-score normalization over $\mathcal{C}_i$. 
For $b \in \{\text{global},\text{local}\}$, the normalized score is:
\vspace{-1pt}
\begin{equation}
\widetilde{S}_b(s_i,v_j)
=
\frac{S_b(s_i,v_j)-\mu_i^{(b)}}{\sigma_i^{(b)}+\epsilon},
\quad v_j \in \mathcal{C}_i,
\label{eq:score_normalization}
\end{equation}
\par
\vspace{-1pt}
\noindent where $\mu_i^{(b)}$ and $\sigma_i^{(b)}$ are the mean and standard deviation of scores from branch $b$ over $\mathcal{C}_i$, and $\epsilon>0$ is a small constant for numerical stability.
Finally, we combine the normalized global and local scores with equal weights:
\vspace{-1pt}
\begin{equation}
S_{\text{final}}(s_i,v_j)
=
\widetilde{S}_{\text{global}}(s_i,v_j)
+
\widetilde{S}_{\text{local}}(s_i,v_j).
\label{eq:final_score}
\end{equation}
\par
\vspace{-1pt}
\noindent DGSC returns the candidates in $\mathcal{C}_i$ in descending order of $S_{\text{final}}$, integrating global identity-level correspondence with fine-grained local matching evidence.

\begin{table*}[t]
    \centering
    \setlength{\abovecaptionskip}{2pt}
    \setlength{\belowcaptionskip}{2pt}
    \caption{\textbf{Comparison with state-of-the-art methods on Speech-PEDES.} All baseline results are reproduced using their publicly available code or checkpoints. Best results are in \textbf{bold}.}
    \label{tab:main}

    \footnotesize
    \renewcommand{\arraystretch}{1.1} 
    
    \begin{tabular}{
        p{0.23\linewidth}|
        >{\centering\arraybackslash}p{0.1\linewidth}|
        >{\centering\arraybackslash}p{0.15\linewidth}
        >{\centering\arraybackslash}p{0.15\linewidth}||cccc
    }
        \hline\thickhline
        \rowcolor{headgray}
        Method & Ref. & Visual Encoder & Audio / Text Encoder & R-1 & R-5 & R-10 & mAP \\
        \hline
        \hline
        \multicolumn{8}{l}{ \textit{General Speech--Image Retrieval} } \\
        \hline
        FaST-VGS$_{\mathrm{CO}}$~\cite{Peng2022FaSTVGS} & TASLP 22 & RN101 & wav2vec 2.0 & 31.09 & 55.39 & 65.95 & 29.99 \\
        FaST-VGS$_{\mathrm{CTF}}$~\cite{Peng2022FaSTVGS} & TASLP 22 & RN101 & wav2vec 2.0 & 33.89 & 58.22 & 68.37 & 32.03 \\
        Parallel SpeechCLIP~\cite{Shih2022SpeechCLIP} & SLT 22 & CLIP-ViT & HuBERT & 12.49 & 27.32 & 36.06 & 12.57 \\
        Hybrid SpeechCLIP+~\cite{Wang2024SpeechCLIPPlus} & ICASSPW 24 & CLIP-ViT & HuBERT & 12.36 & 27.91 & 37.23 & 12.69 \\
        \hline
        \multicolumn{8}{l}{ \textit{Automatic Speech Recognition + Text-Based Person Retrieval (two-stage)} } \\
        \hline
        IRRA~\cite{Jiang2023IRRA} & CVPR 23 & CLIP-ViT & Whisper + CLIP-Text & 56.68 & 76.48 & 83.09 & 52.18 \\
        TBPS-CLIP~\cite{Cao2024TBPSCLIP} & AAAI 24 & CLIP-ViT & Whisper + CLIP-Text & 57.26 & 75.60 & 81.94 & 51.75 \\
        RDE~\cite{Qin2024RDE} & CVPR 24 & CLIP-ViT & Whisper + CLIP-Text & 61.61 & 79.99 & 85.66 & 55.07 \\
        DM-Adapter~\cite{Liu2025DMAdapter} & AAAI 25 & CLIP-ViT & Whisper + CLIP-Text & 57.00 & 75.96 & 83.01 & 51.84 \\
        HCA~\cite{hca} & MM 25 & CLIP-ViT & Whisper + CLIP-Text & 59.67 & 78.15 & 84.37 & 54.83 \\
        GA-DMS~\cite{Zheng2025GADMS} & EMNLP 25 & CLIP-ViT & Whisper + CLIP-Text & 60.92 & 79.27 & 84.86 & 55.02 \\
        MRA~\cite{MRA} & PR 26 & Swin-B & Whisper + BERT & 61.24 & 79.00 & 85.32 & 55.13 \\
        CECA~\cite{Zou2026CECA} & CVPR 26 & CLIP-ViT & Whisper + CLIP-Text & 57.98 & 76.59 & 83.27 & 52.11 \\
        OMRE~\cite{OMRE} & TIFS 26 & CLIP-ViT & Whisper + CLIP-Text & 58.63 & 77.76 & 84.10 & 52.69 \\
        \hline
        \multicolumn{8}{l}{ \textit{Direct Speech-Based Person Retrieval} } \\
        \hline
        \rowcolor[HTML]{D7F6FF}
        \textbf{Pro-SPA (Ours)} & -- & \textbf{ViT} & \textbf{WAVE} & \textbf{63.42} & \textbf{81.90} & \textbf{88.37} & \textbf{55.21} \\
        \hline\thickhline
    \end{tabular}
    \vspace{-14pt}
\end{table*}


The retrieval illustration and core implementation of Pro-SPA are available in our \href{https://anonymous.4open.science/r/SPR-0365/}{anonymous repository}.
Our primary goal is to establish a simple, effective, and reproducible baseline for the novel SPR task.
Pro-SPA incorporates targeted designs to address the challenges of cross-modal speech-image learning: ISA targets decouple identity-discriminative global appearance semantics from appearance-irrelevant acoustic variations, LSSA extracts and aligns sparse local appearance cues in continuous speech without relying on explicit word-level segmentation, and DGSC improves retrieval by integrating these global and local matching cues.
Together with Speech-PEDES, the first dataset for SPR, our work establishes a complete data--model--evaluation pipeline for this emerging direction.

\begin{table}[t]
    \centering
    \setlength{\abovecaptionskip}{3pt}
    \setlength{\belowcaptionskip}{3pt}
    \caption{\textbf{Ablation study of Pro-SPA.} \textbf{Bold} indicates the best performance, and our final configuration is highlighted in blue.}
    \small
    \resizebox{\linewidth}{!}{
        \renewcommand\arraystretch{1.1}
    \begin{tabular}{l|ccc||cccc}
    \hline\thickhline
        \rowcolor{headgray}
        Methods & ISA & LSSA & DGSC & R-1 & R-5 & R-10 & mAP \\
        \hline\hline
        Baseline & & & & 9.29 & 21.93 & 29.99 & 8.90 \\
        $M_1$ & \checkmark & & & 54.01 & 76.04 & 83.74 & 47.13 \\
        $M_2$ & & \checkmark & & 39.38 & 57.54 & 63.94 & 29.02 \\
        $M_3$ & \checkmark & \checkmark & & 57.63 & 78.18 & 85.58 & 49.99 \\
        \hline
        \rowcolor[HTML]{D7F6FF}
        Pro-SPA (Ours) & \checkmark & \checkmark & \checkmark
        & \textbf{63.42} & \textbf{81.90} & \textbf{88.37} & \textbf{55.21} \\
    \hline\thickhline
    \end{tabular}%
    }
    \label{tab:component}
    \vspace{-14pt}
\end{table}

\vspace{-4pt}
\section{Experiments}
\label{sec:formatting}

In this section, we conduct extensive experiments to verify the effectiveness of Pro-SPA. The implementation details and evaluation protocols are in Appendix~\ref{app:implement} and~\ref{app:protocol}.

\vspace{-4pt}
\subsection{Comparison with SOTA Methods}

To evaluate Pro-SPA, we compare it against two representative groups of SOTA methods: \textit{General Speech–Image Retrieval} methods and two-stage \textit{ASR + Text-Based Person Retrieval} pipelines.
The first group performs retrieval by directly applying on Speech-PEDES, whereas the second adopts a cascade strategy that first converts speech into text and then leverages established text-based person retrieval models.
Quantitative results in Tab.~\ref{tab:main} yield three key observations:
\ding{182} Pro-SPA significantly outperforms all general speech-image retrieval methods, achieving substantial gains even over the best competitor (FaST-VGS$_\mathrm{CTF}$). This demonstrates that while models designed for general natural scenes struggle with fine-grained identity discrimination appearance semantics, Pro-SPA successfully extracts and aligns these crucial subtle cues.
\ding{183} Pro-SPA surpasses all two-stage ASR+Text-based Person Retrieval methods. Despite leveraging powerful text-visual encoders trained on large-scale clean data, two-stage models degrade severely during inference due to the vulnerable transcription step. 
In practice, appearance-irrelevant acoustic variations (\emph{e.g.}, timbre, noise) often cause ASR systems to produce transcription errors and semantic omissions, which traditional TBPR models blindly accept and cannot rectify (see Appendix~\ref{app:robutness_to_vary}). 
By establishing an end-to-end alignment from unsegmented speech to images, Pro-SPA bypasses this cascading error propagation.
This indicates that naive two-stage approaches are inadequate for real-world challenges, validating the necessity and value of the proposed direct SPR task.
\ding{184} Achieving SOTA performance across all metrics, Pro-SPA validates the feasibility of direct speech-based person retrieval and provides a simple, effective, and reproducible baseline for future SPR research.

\vspace{-4pt}
\subsection{Ablation Study}
\vspace{-1pt}

\subsubsection{Effectiveness of the Pro-SPA}

Tab.~\ref{tab:component} evaluates the contributions of three key components in Pro-SPA.
Here, the Baseline denotes zero-shot pre-trained WAVE~\cite{Tang2026WAVE}, ranking images via raw feature similarity.
$M_1$ trains ISA and uses $S_\text{global}$ for ranking, whereas $M_2$ trains LSSA without prior ISA training and uses $S_\text{local}$ for ranking. 
$M_3$ adopts our progressive training pipeline but omits DGSC for final ranking refinement.
Finally, Pro-SPA applies DGSC to obtain the final ranking. 
Key observations are:
\ding{182} The poor Baseline performance shows that advanced pre-trained omni-modal models still struggle with the fine-grained SPR task without domain-specific alignment (see Appendix~\ref{app:com_omni}).
\ding{183} $M_1$ and $M_2$ significantly outperform Baseline, proving that learning identity-discriminative appearance semantics across speaker variation (ISA) and explicitly extracting and aligning sparse local cues (LSSA) effectively overcome the core challenges in SPR.
\ding{184} Combining LSSA and ISA ($M_3$) yields further improvements. This confirms that global semantics and local details capture complementary cues. Integrating them enables a more comprehensive cross-modal alignment than relying on either perspective alone.
\ding{185} Pro-SPA achieves the best overall performance,  substantially boosting $M_3$. This highlights DGSC's effectiveness: By dynamically calibrating global and local scores, DGSC maximizes their complementary strengths and achieves optimal cross-modal alignment.

\begin{table}[t]
    \centering
    \setlength{\abovecaptionskip}{2pt}
    \setlength{\belowcaptionskip}{2pt}
    \caption{\textbf{Ablation of LSSA components.} \textbf{Bold} indicates the best performance, and our final configuration is highlighted in blue.}
    \small
    \resizebox{\linewidth}{!}{
        \renewcommand\arraystretch{1.1}
    \begin{tabular}{l|ccc||cccc}
    \hline\thickhline
        \rowcolor{headgray}
        Methods & MSA & BMCA & BCA & R-1 & R-5 & R-10 & mAP \\
        \hline\hline
        Baseline &            &            &            & 54.01 & 76.04 & 83.74 & 47.13 \\
        $M_4$    & \checkmark &            &            & 61.86 & 80.88 & 87.28 & 54.12 \\
        $M_5$    &            & \checkmark &            & 62.30 & 81.32 & 87.93 & 54.74 \\
        $M_6$    & \checkmark & \checkmark &            & 62.39 & 81.60 & 87.95 & 54.84 \\
        $M_7$    &            & \checkmark & \checkmark & 63.03 & 81.64 & 88.27 & 54.98 \\
        \hline
        \rowcolor[HTML]{D7F6FF}
        Pro-SPA (Ours) & \checkmark & \checkmark & \checkmark
        & \textbf{63.42} & \textbf{81.90} & \textbf{88.37} & \textbf{55.21} \\
    \hline\thickhline
    \end{tabular}
    }
    \label{tab:lssa_components}
    \vspace{-12pt}
\end{table}


\begin{table}[t]
    \centering
    \setlength{\abovecaptionskip}{2pt}
    \setlength{\belowcaptionskip}{2pt}
    \caption{\textbf{Contributions of global and local scores.} \textbf{Bold} indicates the best performance, final configuration is highlighted in blue.}
    \footnotesize
        \resizebox{\linewidth}{!}{%
        \renewcommand\arraystretch{1.1}
        \begin{tabular}{l|ccc||cccc}
        \hline\thickhline
        \rowcolor{headgray}
        Scoring Strategy & $S_{\text{global}}$ & $S_{\text{local}}$ & DGSC
        & R-1 & R-5 & R-10 & mAP \\
        \hline\hline
        \emph{Global-only}
        & \checkmark &  &  & 54.01 & 76.04 & 83.74 & 47.13 \\
        \emph{Local only}
        &  & \checkmark &  & 52.99 & 74.74 & 82.57 & 45.98 \\
        \emph{Global + Local}
        & \checkmark & \checkmark & & 57.63 & 78.18 & 85.58 & 49.99 \\
        \hline
        \rowcolor[HTML]{D7F6FF}
        Pro-SPA (Ours)
        & \checkmark & \checkmark & \checkmark & \textbf{63.42} & \textbf{81.90} & \textbf{88.37} & \textbf{55.21} \\
        \hline\thickhline
        \end{tabular}
    }%
    \label{tab:dgsc}
    \vspace{-14pt}
\end{table}


\vspace{-3pt}
\subsubsection{Effectiveness of Local Speech-Segment Alignment}

To extract and align sparse appearance cues without explicit word-level segmentation, we propose the Local Speech-Segment Alignment (LSSA). Tab.~\ref{tab:lssa_components} evaluates its three internal components: Multi-head Self-Attention (MSA), Bidirectional Multi-head Cross-Attention (BMCA), and Bidirectional Consistency Aggregation (BCA). Here, the Baseline corresponds to the ISA-only setting ($M_1$). Key observations are as follows:
\ding{182} Both $M_4$ and $M_5$ substantially outperform the Baseline, validating the effectiveness of MSA and BMCA in extracting and aligning informative features from sparse local cues. Their combination ($M_6$) yields further gains, proving their complementary roles in overcoming the implicit boundaries and variable durations of spoken descriptions.
\ding{183} The consistent improvements from $M_5$ to $M_7$, and from $M_6$ to Pro-SPA, highlight the necessity of BCA. By assigning greater aggregation weights to features with strong reciprocal cross-modal support, BCA effectively filters out appearance-irrelevant noise and adaptively emphasizes the most informative fine-grained cues.
\ding{184} Pro-SPA achieves optimal performance. This confirms that our LSSA, which integrates contextual feature refinement (MSA), bidirectional cross-modal interaction (BMCA), and adaptive consistency weighting (BCA), constitutes a highly effective approach for fine-grained local speech-image alignment.

\vspace{-1pt}
\subsubsection{Contributions of Global and Local Scores}

Tab.~\ref{tab:dgsc} evaluates the contributions of the global ($S_{\text{global}}$) and local ($S_{\text{local}}$) scores during inference, as well as the effects of score fusion strategies.
\emph{Global-only} and \emph{Local-only} rank candidate images using only $S_{\text{global}}$ and $S_{\text{local}}$, respectively, while \emph{Global + Local} directly sums the two scores, and Pro-SPA combines them via Dual-Granularity Score Calibration (DGSC). 
Key observations are as follows:
\ding{182} Combining both scores outperforms using either alone, demonstrating that global identity-level matching (ISA) and fine-grained local matching (LSSA) capture complementary cues.
\ding{183} DGSC further improves performance, validating the effectiveness of DGSC in calibrating and integrating global and local match scores to optimize the final retrieval ranking.

\vspace{-3pt}
\subsection{Visualization of Retrieval Ranking}

\begin{figure}[t]
    \centering
    \setlength{\abovecaptionskip}{4pt}
    \setlength{\belowcaptionskip}{4pt}
    \includegraphics[width=\linewidth]{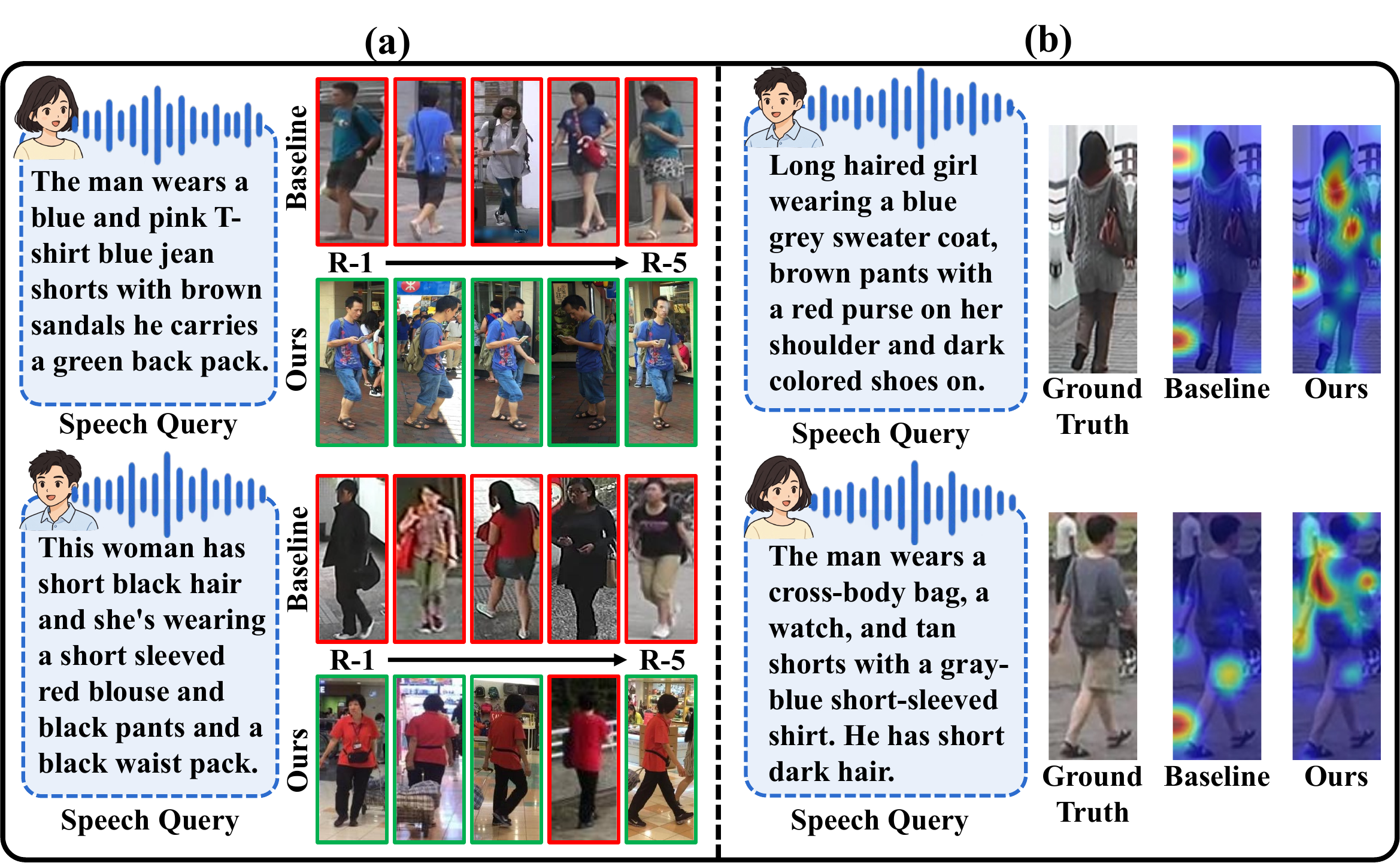}
    \caption{\textbf{Visualization of retrieval rankings and cross-modal attention maps.} We compare results of Baseline and our Pro-SPA.}
    \label{fig:visual_rank_and_att}
    \vspace{-15pt}
\end{figure}

We visualize the retrieval rankings and cross-modal attention maps in Fig.~\ref{fig:visual_rank_and_att}. 
Fig.~\ref{fig:visual_rank_and_att}~(a) shows that Pro-SPA significantly improves retrieval performance over the baseline.
Furthermore, the attention maps in Fig.~\ref{fig:visual_rank_and_att}~(b) reveal that Pro-SPA focuses on highly identity-discriminative regions, whereas the baseline rarely attends to these appearance features. This confirms Pro-SPA is capable to localize specific visual regions described in continuous speech. 
Together, these results demonstrate that Pro-SPA effectively learns discriminative speech representations and aligns sparse appearance cues without explicit word-level segmentation.



Additional details, visualizations, and experiments (\emph{e.g.}, dataset statistics, retrieval and web demos, robustness tests, inference efficiency, ASR baselines, and further ablations) are provided in the Appendix.

\vspace{-4pt}
\section{Conclusion}
\label{sec:conclusion}
\vspace{-2pt}

Recognizing the inherent limitations of traditional TBPR where well-formed textual descriptions are not always readily available in real-world scenarios, we introduce Speech-based Person Retrieval (SPR), a more natural and effective paradigm for person retrieval.
For this, we build the first benchmark Speech-PEDES, combining authentic human recordings and highly realistic synthesized speech. 
To tackle this task, we proposed Pro-SPA, which progressively learns identity-discriminative speech representations and aligns sparse appearance cues across modalities without explicit word boundaries. 
Extensive experiments demonstrate the potential of this direct speech-driven paradigm in person retrieval and effectiveness of our method. 
Together, Speech-PEDES and Pro-SPA establish a complete data--model--evaluation pipeline, providing a simple, effective, and reproducible baseline for future SPR research.

\section*{LLM Usage Statement}

For manuscript preparation, we used generative AI tools to aid writing and polish the language, including grammar, wording, and readability. The authors reviewed and revised all AI-assisted text to ensure that it preserved the intended technical meaning and accurately reflected the methods and results presented in the paper.
The authors take full responsibility for the final content of this work, including the text, scientific claims, experimental results, and artifacts produced with the assistance of generative AI.

{
    \small
    \bibliographystyle{ieeenat_fullname}
    \bibliography{main}
}

\clearpage
\setcounter{page}{1}
\maketitlesupplementary
\setcounter{section}{6}
\setcounter{figure}{5}     
\setcounter{table}{5}

\makeatletter
\newcommand{\suppnavsection}[2]{%
  \vspace{6pt}%
  \@dottedtocline{1}{0em}{1.9em}%
    {\hyperref[#1]{\bfseries\numberline{\ref*{#1}.}#2}}%
    {\hyperref[#1]{\bfseries\pageref*{#1}}}}
\newcommand{\suppnavsubsection}[2]{%
  \@dottedtocline{2}{0.65em}{2.8em}%
    {\hyperref[#1]{\numberline{\ref*{#1}}#2}}%
    {\hyperref[#1]{\pageref*{#1}}}}
\newcommand{\suppnavsubsubsection}[2]{%
  \@dottedtocline{3}{1.3em}{3.7em}%
    {\hyperref[#1]{\numberline{\ref*{#1}}#2}}%
    {\hyperref[#1]{\pageref*{#1}}}}
\makeatother

\section*{Contents}
\suppnavsection{app:discussion_statements}{Discussion and Statements}
\suppnavsubsection{app:ethics}{Ethics Statement}
\suppnavsubsection{app:reproducibility}{Reproducibility Statement}
\suppnavsubsection{app:limitations}{Limitations and Future Work}
\suppnavsection{sec:notation}{Notation Summary}
\suppnavsection{app:experiment_settings}{Experimental Settings}
\suppnavsubsection{app:implement}{Implementation Details}
\suppnavsubsection{app:protocol}{Evaluation Protocol}
\suppnavsection{app:prompts}{Person-Centric Prompts}
\suppnavsection{app:dataset}{Speech-PEDES Dataset Details}
\suppnavsubsection{app:data_split}{Dataset Split}
\suppnavsubsection{app:data_construction}{Data Construction Details}
\suppnavsubsubsection{app:recording_platform}{Web-Based Recording Platform}
\suppnavsubsubsection{app:speech_synthesis}{Speech Perturbation}
\suppnavsubsection{app:data_assessment}{Additional Analysis of Speech-PEDES}
\suppnavsubsubsection{app:speech_statistics}{Utterance Duration and Speaking Rate}
\suppnavsubsubsection{app:speech_quality_rate}{Speech Quality Across Speaking Rate}
\suppnavsubsubsection{app:attribute_errors}{Attribute Recognition Errors}
\suppnavsubsubsection{app:tts_comparison}{Comparison of Text To Speech Models}
\suppnavsubsection{app:dataset_examples}{Examples of Speech-Image Pairs}

\suppnavsection{app:additional_experiments}{Additional Experiments and Analyses}
\suppnavsubsection{app:com_omni}{Comparison with SOTA Omni-modal Models}
\suppnavsubsection{app:asr_models}{Effect of ASR Models}
\suppnavsubsection{app:contrastive_objectives}{Comparison of Contrastive Objectives}
\suppnavsubsection{app:speech_segments}{Hyper-Parameter Analysis}
\suppnavsubsection{app:inference_efficiency}{Inference Efficiency}
\suppnavsubsection{app:robutness_to_vary}{Robustness to Speech Variations}
\suppnavsubsubsection{app:asr_noise}{ASR Transcription Errors}
\suppnavsubsubsection{app:noise_robustness}{Robustness Against Background Noise}
\suppnavsubsubsection{app:robust_to_speaker}{Robustness to Speaker Variations}
\suppnavsubsubsection{app:icc_speakers}{Speaker Robustness of ICC vs.\ InfoNCE}
\suppnavsubsection{app:hn_effect_lssa}{Effect of negative sampling strategies in LSSA}
\suppnavsubsection{app:compare_agg_lssa}{Effect of aggregation strategies in LSSA}
\suppnavsection{app:visualization}{Visualizations}
\suppnavsubsection{app:attention_visualization}{Attention Visualization}
\suppnavsubsubsection{app:local_alignment}{Speech Attention: Pro-SPA vs.\ Baseline}
\suppnavsubsubsection{app:visualization_info_vs_ours}{Cross-Modal Attention: ICC vs.\ InfoNCE}
\suppnavsubsection{app:visualization_of_rank}{Visualization of Retrieval Rankings}
\suppnavsubsection{app:visualization_retrieval}{Visualization of Retrieval Process}

\section{Discussion and Statements}
\label{app:discussion_statements}

\subsection{Ethics Statement}
\label{app:ethics}

The Speech-PEDES dataset comprises person images and textual captions sourced from CUHK-PEDES~\cite{Li_2017_CVPR}, paired with authentic human speech recordings and synthesized speech generated via reference voices from GLOBE~\cite{globe}. 
All human participants who contributed to the real speech recordings provided explicit informed consent for their voice data to be utilized for academic research and included in the public dataset release. 
Speech-PEDES is strictly intended to advance research in speech-based person retrieval. 
We strongly advocate for its responsible use in compliance with ethical guidelines and individual privacy protection.

\subsection{Reproducibility Statement} 
\label{app:reproducibility}

To support reproducibility and provide intuitive demonstrations, we release the core implementation of our Pro-SPA framework and our web-based recording platform in our \href{https://anonymous.4open.science/r/SPR-0365/}{anonymous repository}. 
In addition to the core source code, the repository includes playable speech samples, a visual walkthrough of the retrieval process, and qualitative retrieval result demos. 
Detailed descriptions and interface screenshots of the data collection platform are provided in Appendix~\ref{app:recording_platform}, while visual illustrations of the speech samples and the retrieval process are presented in Appendices~\ref{app:dataset_examples} and~\ref{app:visualization_retrieval}.

\subsection{Limitations and Future Work}
\label{app:limitations}

Speech-PEDES combines authentic human recordings with synthesized speech featuring varied speaking styles.
Although these data capture practical variations, including pauses and changes in speaking rate, both the recordings and synthesized utterances are based on predefined descriptions.
The dataset therefore may not fully reflect the linguistic variability of spontaneous descriptions (\emph{e.g.}, self-corrections, incomplete expressions, and freely formulated content).
In addition, the overall speech quality in Speech-PEDES remains relatively high, limiting its coverage of challenging acoustic conditions in complex real-world environments (\emph{e.g.}, heavy background noise, multi-talker overlapping speech, or far-field reverberation).
Collecting spontaneous spoken descriptions and evaluating Pro-SPA under more challenging real-world audio conditions remain important directions for future work.

Nevertheless, this study serves as a foundational step toward Speech-based Person Retrieval (SPR), focusing on validating the feasibility and necessity of this practical yet underexplored paradigm beyond classical ReID and TBPR.
To this end, we establish a reliable evaluation cornerstone with Speech-PEDES and propose Pro-SPA, a simple, effective, and reproducible baseline that enables direct fine-grained alignment from unsegmented speech.
Moreover, our analyses (see Appendix~\ref{app:robutness_to_vary} for details) show that Pro-SPA already exhibits robustness against simulated noise and speaker variations.
We hope our work, together with Speech-PEDES and Pro-SPA, will establish a complete data--model--evaluation pipeline, serving as a solid foundation to facilitate further research in this emerging field and help extend SPR to unconstrained real-world scenarios.

\section{Notation Summary}
\label{sec:notation}

In Tab.~\ref{tab:notation_prospa}, we present a summary of the key notations and variables used in the formulation of our Pro-SPA.

\begin{table}[t]
    \centering
    \setlength{\abovecaptionskip}{3pt}
    \setlength{\belowcaptionskip}{3pt}
    \setlength{\tabcolsep}{3pt}
    \small
    \caption{Summary of key notations used in Pro-SPA.}
    \resizebox{\linewidth}{!}{
    \renewcommand\arraystretch{1.2}
    \begin{tabular}{l||lc}
        \hline\thickhline
        \rowcolor{headgray}
        Symbol & Description & Reference \\
        \hline
        $s_i,\ v_j$
        & Speech query and image
        & Eq.\ref{eq:global_similarity} \\
        $K_h,\ K_g$
        & Mining and reranking sizes
        & Secs.~\ref{sec:lssa},~\ref{sec:dgsc} \\
        \hline
        $g_i^s,\ g_j^v$
        & Global embeddings
        & Eq.\ref{eq:global_embeddings} \\
        $S_{\text{global}},\ S_{i,j}$
        & Global matching score
        & Eq.\ref{eq:global_similarity} \\
        $P_i,\ P_j$
        & Same-identity positive sets
        & Eq.\ref{eq:icc_s2v},~Eq.\ref{eq:icc_v2s} \\
        $B,\ \tau$
        & Batch size and ICC temperature
        & Eq.\ref{eq:icc_s2v} \\
        $\mathcal{L}_{\text{ICC}}$
        & Identity-aware contrastive loss
        & Eq.\ref{eq:icc_loss} \\
        \hline
        $m,\ n,\ d$
        & Segment/region counts and dimension
        & Eq.\ref{eq:local_projection} \\
        $F^s,\ F^v$
        & Projected local features
        & Eq.\ref{eq:local_projection} \\
        $\bar{F}^s,\ \bar{F}^v$
        & Refined local features
        & Eq.\ref{eq:local_refinement} \\
        $W^s,\ W^v,\ W^r$
        & Attention and consistency maps
        & Eq.\ref{eq:bidirectional_consistency} \\
        $\alpha_k,\ \beta_l,\ Z$
        & Aggregation weights and normalizer
        & Eq.\ref{eq:aggregation_weights} \\
        $\hat{F}^s,\ \hat{F}^v$
        & Aggregated local features
        & Eq.\ref{eq:local_aggregation} \\
        $S_{\text{local}}$
        & Local matching score
        & Eq.\ref{eq:local_similarity} \\
        $v_i^+,\ \tilde{N}_i$
        & Positive image and candidate set
        & Eq.\ref{eq:lssa_loss} \\
        $\tau_l,\ \mathcal{L}_{\text{LSSA}}$
        & Local temperature and loss
        & Eq.\ref{eq:lssa_loss} \\
        \hline
        $\mathcal{C}_i$
        & Reranking candidate set
        & Eq.\ref{eq:score_normalization} \\
        $\mu_i^{(b)},\ \sigma_i^{(b)},\ \epsilon$
        & Mean, std., and stability constant
        & Eq.\ref{eq:score_normalization} \\
        $S_b,\ \widetilde{S}_b$
        & Original and normalized scores
        & Eq.\ref{eq:score_normalization} \\
        $S_{\text{final}}$
        & Final reranking score
        & Eq.\ref{eq:final_score} \\
        \hline\thickhline
    \end{tabular}}
    \label{tab:notation_prospa}
    \vspace{-12pt}
\end{table}

\begin{table}[t]
    \centering
    \setlength{\abovecaptionskip}{3pt}
    \setlength{\belowcaptionskip}{3pt}
    \caption{
        Training settings and model hyperparameters of ISA and LSSA.
        Stage-2 settings refer to LSSA, while ISA remains frozen.
    }
    \small
    \resizebox{\linewidth}{!}{
        \renewcommand{\arraystretch}{1.15}
        \begin{tabular}{l||c|c}
            \hline\thickhline
            \rowcolor{headgray}
            Configuration & ISA (Stage 1) & LSSA (Stage 2) \\
            \hline\hline

            epochs                      & 5             & 10 \\
            total batch size            & 32            & 128 \\
            image resolution            & -             & $224\times224$ \\
            optimizer                   & AdamW~\cite{adamw} & AdamW~\cite{adamw} \\
            optimizer $\beta_1,\beta_2$ & 0.9, 0.999 & 0.9, 0.999 \\
            weight decay                & 0 & $10^{-4}$ \\
            learning rate               & $2\times10^{-5}$ & $10^{-4}$ \\
            warmup steps                & 160 & 0 \\
            temperature                 & 0.01 & 0.07 \\
            numerical precision         & bf16 & bf16 AMP \\
            GPUs for training           & $8\times$A40 & $2\times$A40 \\
            LoRA rank                   & 128 & - \\
            LoRA scaling factor $\alpha$ & 256 & - \\
            LoRA dropout                & 0.05 & - \\
            attention heads in MSA/BMCA & - & 8 \\
            attention dropout in MSA/BMCA & - & 0.1 \\

            \hline\thickhline
        \end{tabular}
    }
    \label{tab:supp_impl}
    \vspace{-12pt}
\end{table}

\section{Experimental Settings}
\label{app:experiment_settings}

\subsection{Implementation Details}
\label{app:implement}


We adopt WAVE-7B~\cite{Tang2026WAVE} as our backbone. Speech inputs are resampled to 16\,kHz and converted into 128-channel log-Mel spectrograms.
We train ISA with $\mathcal{L}_{\mathrm{ICC}}$ by updating only the LoRA~\cite{lora} parameters in the query, key, and value projections of each LLM layer.
ISA is then frozen while LSSA is trained from scratch.
For LSSA, final-layer speech hidden states are temporally average-pooled into 32 tokens, while images are resized to $224\times224$, yielding 64 visual tokens. Both token sets are projected to $d=512$ dimensions. LSSA training combines the top-128 hard negatives with in-batch negatives.
During inference, LSSA re-scores the top-$K_g$ ($K_g=150$) candidates retrieved by ISA, and DGSC fuses their z-score-normalized global and local scores.
All of our experiments are conducted on NVIDIA A40 GPUs.
Additional hyperparameters and training configurations are provided in Tab.~\ref{tab:supp_impl}.

\subsection{Evaluation Protocol}
\label{app:protocol}
We adopt the widely-used Rank-K (R-K for short, K=$1, 5, 10$) metric in our experiments, which denotes the percentage of successful retrieval that the ground-truth image is found within the top K ranked results.
In addition, we also adopt the mean average precision (mAP) as a complementary metric.
Rank-K reflects the accuracy of the first few retrieval results, while mAP emphasizes the comprehensive performance of the method.

\section{Person-Centric Prompts}
\label{app:prompts}

We use person-centric system messages $T_s$ and $T_v$ to guide WAVE's shared Thinker toward pedestrian appearance semantics in speech and images, respectively. Each message is appended to the corresponding modality tokens before the \texttt{[EOS]} token. The exact messages are provided below.

\begin{center}
    \begin{tcolorbox}[
        colback=gray!5,      
        colframe=gray!50,    
        boxrule=1pt,       
        arc=3pt,             
        width=0.98\linewidth, 
        left=6pt, right=6pt, top=6pt, bottom=6pt]
        {\small\bfseries
        Person-centric system message for speech encoder ($T_s$):\par}
        \smallskip
        {\small\ttfamily
        Represent the visible appearance of the pedestrian described in the speech.\par}
    \end{tcolorbox}
\end{center}

\begin{center}
    \begin{tcolorbox}[
        colback=gray!5,      
        colframe=gray!50,    
        boxrule=1pt,       
        arc=3pt,             
        width=0.98\linewidth, 
        left=6pt, right=6pt, top=6pt, bottom=6pt]
        {\small\bfseries
        Person-centric system message for image encoder ($T_v$):\par}
        \smallskip
        {\small\ttfamily
        Represent the visible appearance of the pedestrian described in the image.\par}
    \end{tcolorbox}
\end{center}

\begin{figure*}[t]
    \centering
    \setlength{\abovecaptionskip}{3pt}
    \setlength{\belowcaptionskip}{3pt}
    \includegraphics[width=0.95\linewidth]{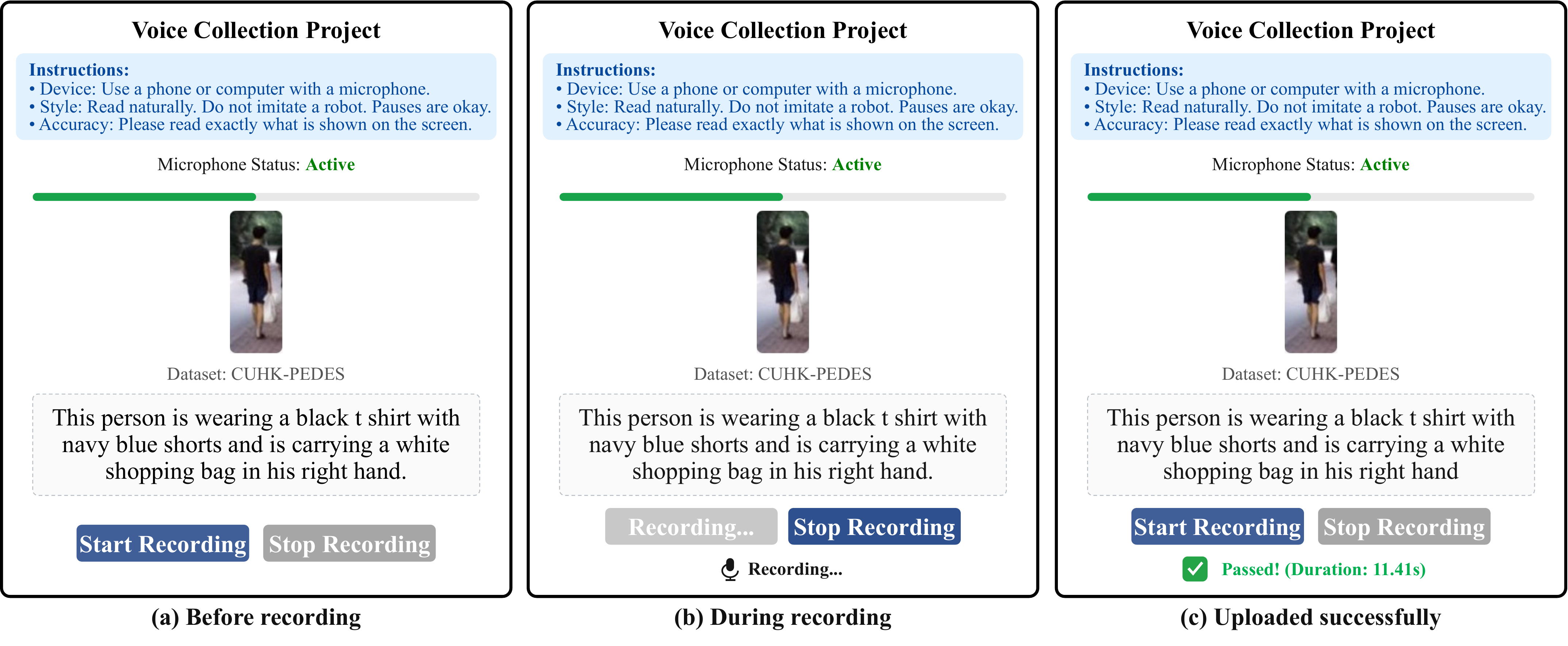}
    \caption{\textbf{Recording and submission interfaces of the web-based platform.} (a) Prompt presentation before recording; (b) speech recording in progress; and (c) upload confirmation after passing HSCV verification.}
    \label{fig:supp_web_demo1}
    \vspace{-10pt}
\end{figure*}

\begin{figure*}[t]
    \centering
    \setlength{\abovecaptionskip}{3pt}
    \setlength{\belowcaptionskip}{3pt}
    \includegraphics[width=0.95\linewidth]{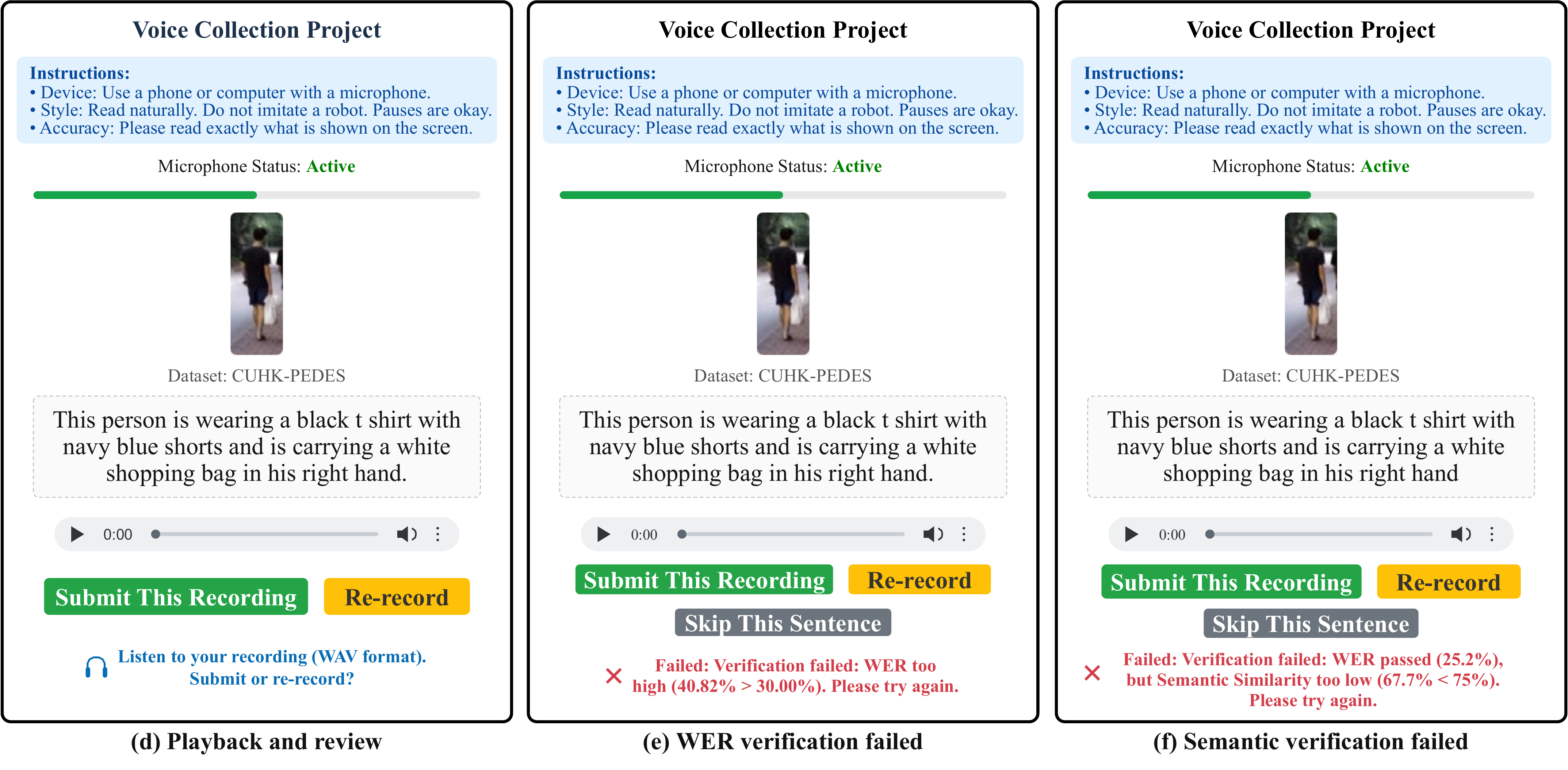}
    \caption{\textbf{Playback review and rejection feedback on the web-based platform.} (d) Playback review with submit and re-record options; (e) rejection due to excessive WER; and (f) rejection due to insufficient semantic similarity within the intermediate WER range.}
    \label{fig:supp_web_demo2}
    \vspace{-10pt}
\end{figure*}

\section{Speech-PEDES Dataset Details}
\label{app:dataset}

\subsection{Dataset Split}
\label{app:data_split}
Speech-PEDES comprises 37,128 images across 12,003 pedestrian identities, accompanied by a total of 74,282 speech queries. 
Following the data split protocol of CUHK-PEDES~\cite{Li_2017_CVPR}, we allocate 34,054 images from 11,003 identities (paired with 68,126 speech queries) to the training set. 
The remaining 3,074 images from 1,000 identities (paired with 6,156 speech queries) form the test set. 
Notably, the test set consists exclusively of human-recorded speech to ensure a realistic and reliable evaluation.

\subsection{Data Construction Details}
\label{app:data_construction}

\subsubsection{Web-Based Recording Platform}
\label{app:recording_platform}


We develop a web-based recording platform that integrates task presentation, speech recording, playback verification, and quality control into a unified interface. 
During collection, the platform displays the pedestrian image and its corresponding reference description, accompanied by reading guidelines and microphone status indicators (Fig.~\ref{fig:supp_web_demo1}~(a)). 
Participants click to start recording, read the text in a natural tone, and stop after completion (Fig.~\ref{fig:supp_web_demo1}~(b)).
Afterward, they can play back the speech to check content and quality, and then submit or re-record (Fig.~\ref{fig:supp_web_demo2}~(d)).

Submitted recordings are validated by the Hierarchical Speech Content Verification (HSCV) mechanism, which directly accepts, rejects, or triggers secondary semantic verification based on the Word Error Rate (WER) results. 
Once accepted and uploaded, the interface displays a confirmation message along with the duration of the speech (Fig.~\ref{fig:supp_web_demo1}(c)). 
For rejected recordings, the platform specifies the exact reason: a WER exceeding the threshold of 30\% (Fig.~\ref{fig:supp_web_demo2}(e)) or a semantic similarity below 75\% (Fig.~\ref{fig:supp_web_demo2}(f)). 
In both cases, participants can either re-record or skip the current sentence. Combining manual playback checks with automated feedback, this workflow enables participants to efficiently review and refine their recordings.

\begin{figure}[t]
    \centering
    \setlength{\abovecaptionskip}{3pt}
    \setlength{\belowcaptionskip}{3pt}
    \includegraphics[width=\linewidth]{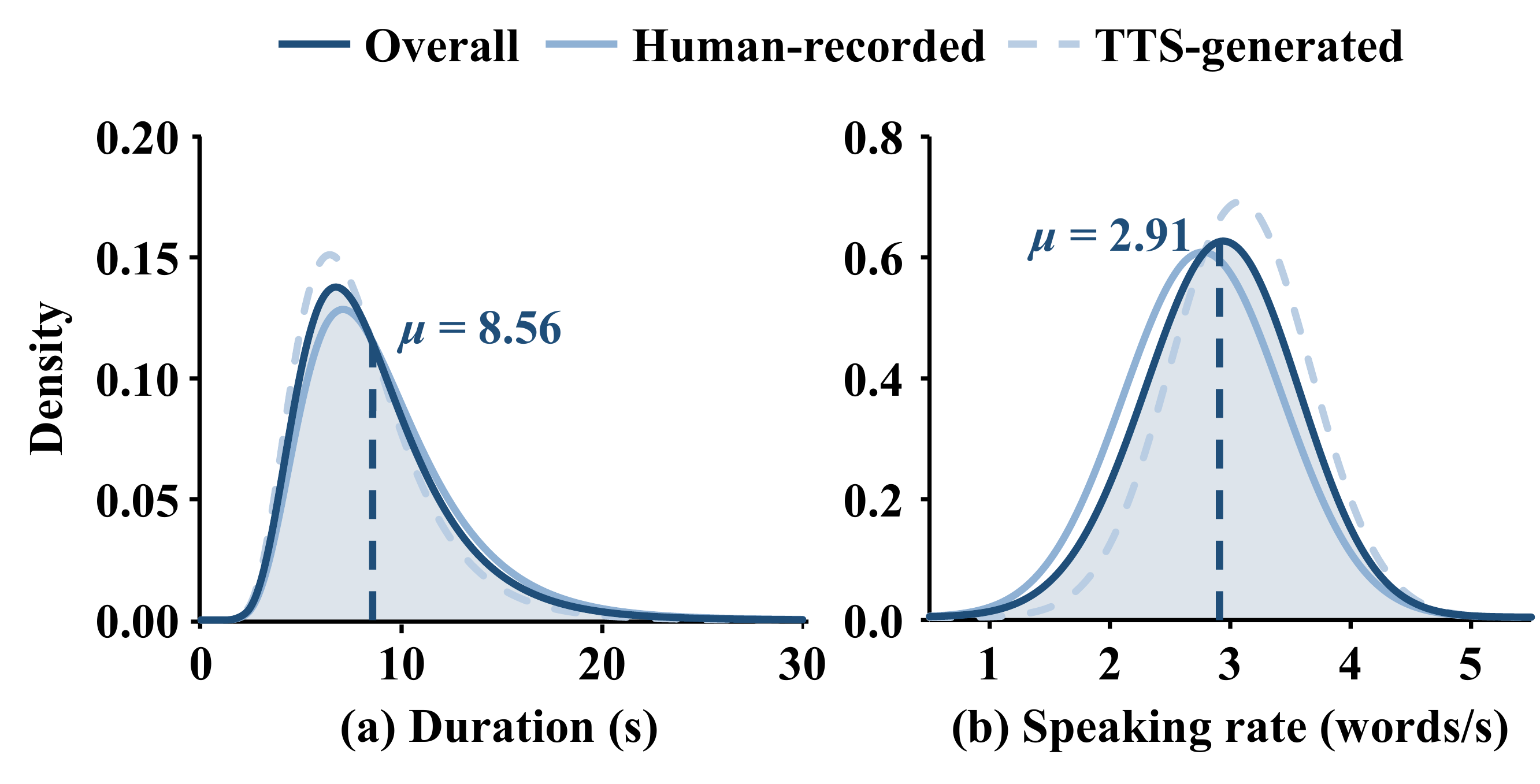}
    \caption{Density distributions of utterance duration and speaking rate on Speech-PEDES. Comparison across the human-recorded, TTS-generated, and overall speech subsets.}
    \label{fig:supp_duration}
    \vspace{-10pt}
\end{figure}

\begin{figure}[t]
    \centering
    \setlength{\abovecaptionskip}{3pt}
    \setlength{\belowcaptionskip}{3pt}
    \includegraphics[width=\linewidth]{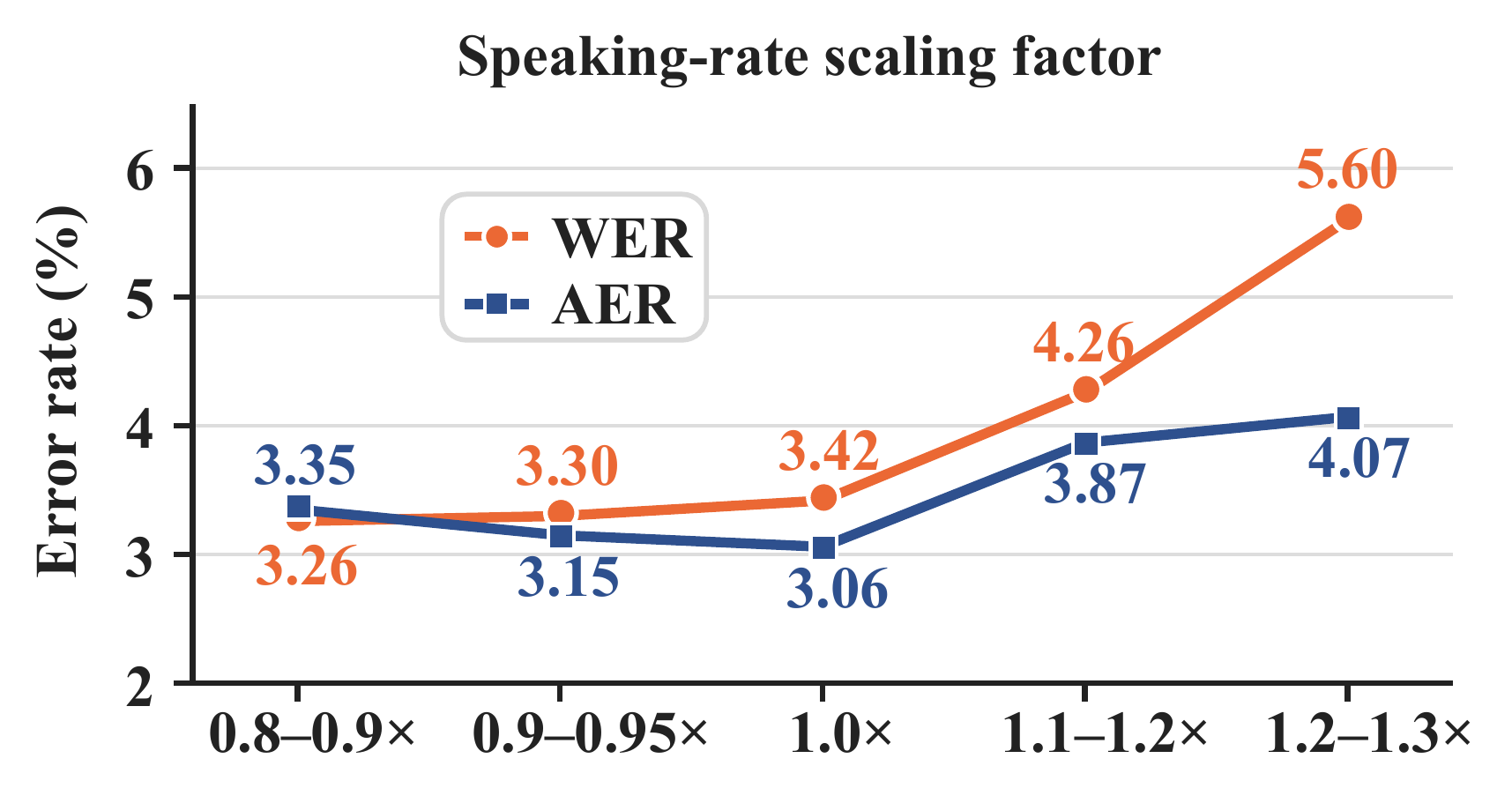}
    \caption{Distribution of Word Error Rate (WER) and Attribute Error Rate (AER) across varying speaking rates on Speech-PEDES.}
    \label{fig:supp_rate_error}
    \vspace{-10pt}
\end{figure}

\begin{figure}[t]
    \centering
    \setlength{\abovecaptionskip}{3pt}
    \setlength{\belowcaptionskip}{3pt}
    \includegraphics[width=\linewidth]{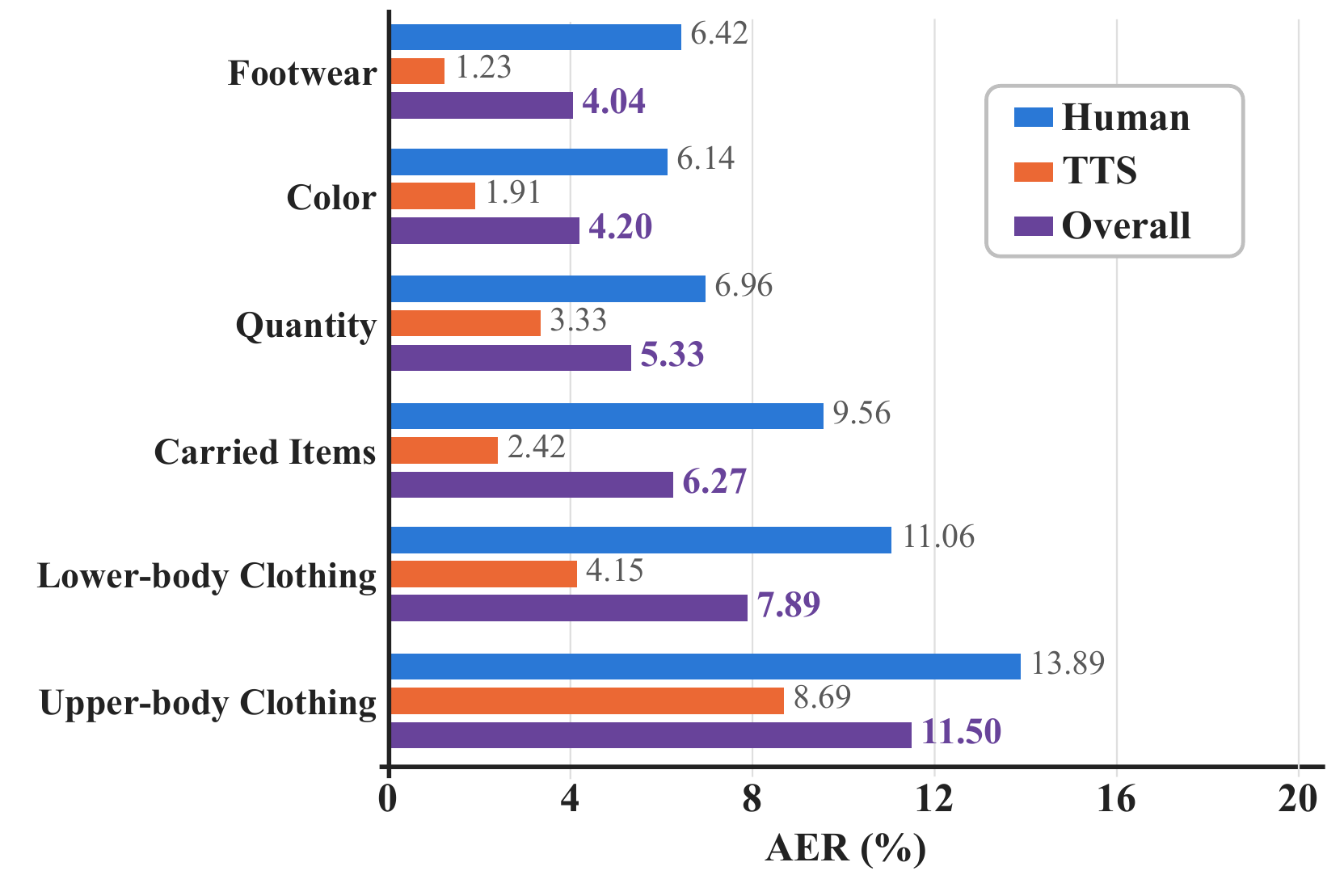}
    \caption{\textbf{Attribute Error Rate (AER) on Speech-PEDES.} We compare the AER across six pedestrian attribute categories for the human-recorded, TTS-generated, and overall subsets.}
    \label{fig:supp_aer}
    \vspace{-10pt}
\end{figure}

\subsubsection{Speech Perturbation}
\label{app:speech_synthesis}

We randomly select 20\% of the reference descriptions and insert spoken filler words (\emph{e.g.}, \textit{uh}, \textit{umm}, \textit{hmm}, \textit{I mean}, and \textit{like}) so that the synthesized speech more closely resembles natural conversational speech. Insertion positions are randomly sampled while avoiding the beginning and end of sentences, with a preference for the first third and middle portions.
In addition, we apply random speed perturbation to the synthesized speech to simulate natural variations in speaking rate. Specifically, 15\% of the samples are scaled to $1.1\text{--}1.2\times$ the original speed, 15\% to $0.9\text{--}0.95\times$, 5\% to $1.2\text{--}1.3\times$, and another 5\% to $0.8\text{--}0.9\times$.

\subsection{Additional Analysis of Speech-PEDES}
\label{app:data_assessment}


\subsubsection{Utterance Duration and Speaking Rate}
\label{app:speech_statistics}

Fig.~\ref{fig:supp_duration} visualizes the density distributions of both speech duration and speaking rate. For both metrics, the distributions of human-recorded and TTS-generated speech exhibit a high degree of consistency. This close overlap confirms the similarity in temporal characteristics between the synthesized and authentic speech data.

\subsubsection{Speech Quality Across Speaking Rate}
\label{app:speech_quality_rate}

Fig.~\ref{fig:supp_rate_error} evaluates the Word Error Rate (WER) and Attribute Error Rate (AER) of the synthesized speech under varying speaking rates. The results show that the error rates remain relatively low across the evaluated speaking rates within our dataset. This demonstrates that our Speech-PEDES dataset effectively expands prosodic diversity to reflect real-world speaking variations, while maintaining reliable fidelity to the original textual content.

\subsubsection{Attribute Recognition Errors}
\label{app:attribute_errors}
Fig.~\ref{fig:supp_aer} details the Attribute Error Rate (AER) across six fine-grained pedestrian attributes. Across all categories, TTS-generated speech exhibits consistently low AERs, reflecting clear pronunciation and accurate attribute preservation in the synthesized subset. Although authentic human speech introduces slightly higher AERs due to real-world recording conditions and accent variations, the overall dataset still maintains relatively low attribute error rates across all categories, confirming the attribute-level reliability of Speech-PEDES.

\begin{figure*}[t]
    \centering
    \setlength{\abovecaptionskip}{3pt}
    \setlength{\belowcaptionskip}{3pt}
    \includegraphics[width=0.95\linewidth]{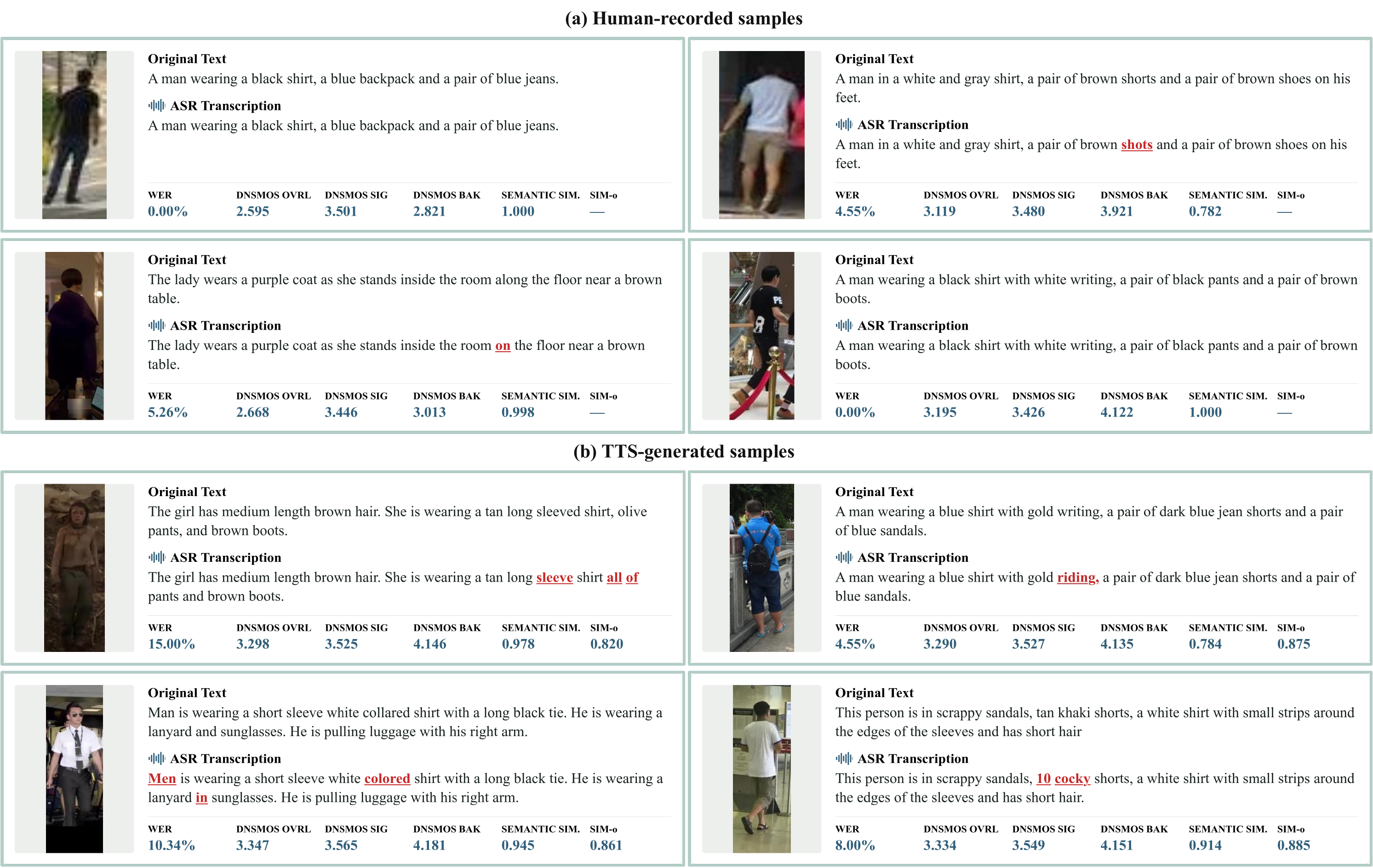}
    \caption{\textbf{Representative examples from Speech-PEDES.} (a) Human-recorded samples and (b) TTS-generated samples.}
    \label{fig:speech_image_pairs}
    \vspace{-12pt}
\end{figure*}

\subsubsection{Comparison of Text To Speech Models}
\label{app:tts_comparison}

Tab.~\ref{tab:supp_tts} presents a comprehensive comparison of candidate Text To Speech (TTS) models evaluated for the construction of our data set. 
F5-TTS~\cite{f5-tts} achieves the lowest Word Error Rate (WER: 3.59\%) and Attribute Error Rate (AER: 3.23\%), along with the highest semantic similarity (96.30\%), while maintaining competitive speaker similarity (SIM-o of 0.87). 
Consequently, these quality evaluation results support selecting F5-TTS to generate the synthesized speech subset of the Speech-PEDES benchmark.

\subsection{Examples of Speech-Image Pairs}
\label{app:dataset_examples}


Fig.~\ref{fig:speech_image_pairs} presents representative examples from Speech-PEDES, comprising four human-recorded and four TTS-generated samples. 
Each example pairs a pedestrian image with its original text description and the ASR transcription of the speech query, where misrecognized words are highlighted in red. 
We also report the corresponding quality metrics: WER, DNSMOS (OVRL/SIG/BAK), semantic similarity, and speaker similarity (SIM-o, for TTS samples only). 
Since speech cannot be directly presented in a static figure, playable speech files of these samples and a video walkthrough are provided in our \href{https://anonymous.4open.science/r/SPR-0365/}{anonymous repository}.

\vspace{-3pt}
\section{Additional Experiments and Analyses}
\label{app:additional_experiments}

\subsection{Comparison with SOTA Omni-modal Models}
\label{app:com_omni}


We evaluate several recent SOTA omni-modal foundation models on our Speech-PEDES under a zero-shot setting. As shown in Tab.~\ref{tab:supp_omni}, despite their extensive multi-modal pre-training, these generalist models exhibit sub-optimal performance on our benchmark. 
Even the strongest zero-shot competitor, e5-omni-7B, achieves only 27.21\% in R-1, lagging far behind our method. 
These results strongly corroborate our claim: advanced pre-trained omni-modal models still struggle with the fine-grained SPR task without domain-specific alignment. 
In contrast, Pro-SPA significantly outperforms all these models, demonstrating its efficacy in overcoming the inherent challenges in SPR task.

\begin{table}[t]
    \centering
    \setlength{\abovecaptionskip}{3pt}
    \setlength{\belowcaptionskip}{3pt}
    \caption{\textbf{Comparison of candidate TTS models for Speech-PEDES construction.} WER, AER, and Sem.\ Sim.\ denote Word Error Rate, Attribute Error Rate, and semantic similarity, respectively. DNSMOS P.835 (OVRL, SIG, BAK) evaluates overall, signal, and background noise quality, while SIM-o measures objective speaker similarity. Best results are in \textbf{bold}, and our selected model is highlighted in blue.}
    \label{tab:supp_tts}
    \footnotesize
    \setlength{\tabcolsep}{5pt}
    \renewcommand{\arraystretch}{1.15}
    \resizebox{\linewidth}{!}{
    \begin{tabular}{l||>{\columncolor[HTML]{D7F6FF}}ccc}
        \hline\thickhline
        \rowcolor{headgray}
        Metric & F5-TTS~\cite{f5-tts} & CosyVoice2~\cite{cosyvoice2} & GPT-SoVITS v3~\cite{gptsovits} \\
        \hline\hline
        WER~$\downarrow$          & \textbf{3.59\%}  & 5.49\%  & 11.67\% \\
        AER~$\downarrow$          & \textbf{3.23\%}  & 4.40\%  & 10.55\% \\
        Sem.\ Sim.~$\uparrow$     & \textbf{96.30\%} & 94.67\% & 90.65\% \\
        \hline
        DNSMOS-OVRL~$\uparrow$    & 3.79 & \textbf{3.95} & 3.88 \\
        DNSMOS-SIG~$\uparrow$     & 3.98 & \textbf{4.12} & 4.09 \\
        DNSMOS-BAK~$\uparrow$     & 4.39 & \textbf{4.52} & 4.42 \\
        \hline
        SIM-o~$\uparrow$          & 0.87 & 0.86 & \textbf{0.88} \\
        \hline\thickhline
    \end{tabular}
    }
    \vspace{-12pt}
\end{table}

\begin{table}[t]
  \centering
  \footnotesize
  \setlength{\abovecaptionskip}{3pt}
  \setlength{\belowcaptionskip}{3pt}
  \setlength{\tabcolsep}{4pt}
  \caption{\textbf{Comparison with state-of-the-art omni-modal models on Speech-PEDES}. All omni-modal models are evaluated under zero-shot setting, while Pro-SPA is trained on Speech-PEDES. Best results are in \textbf{bold}.}
  \begin{tabular}{l||cccc}
    \hline\thickhline
    \rowcolor{headgray}
    Methods & R-1 & R-5 & R-10 & mAP \\
    \hline
    WAVE-7B~\cite{Tang2026WAVE}   &  9.29 & 21.93 & 29.99 &  8.90 \\
    Omni-Embed-Nemotron-3B~\cite{Nemotron}  & 11.39 & 23.13 & 30.25 &  9.75 \\
    OmniEmbed-v0.1~\cite{Tevatron}  & 16.37 & 32.36 & 40.32 & 13.73 \\
    LCO-EMB-Omni (7B)~\cite{LCO}       & 16.73 & 32.67 & 41.99 & 14.30 \\
    e5-omni-7B~\cite{E5-Omni}   & 27.21 & 46.73 & 55.64 & 22.90 \\
    \hline
    \rowcolor[HTML]{D7F6FF}
    Pro-SPA (Ours)  & \textbf{63.42} & \textbf{81.90} & \textbf{88.37} & \textbf{55.21} \\
    \hline\thickhline
  \end{tabular}
  \label{tab:supp_omni}
    \vspace{-10pt}
\end{table}

\vspace{-3pt}
\subsection{Effect of ASR Models}
\label{app:asr_models}

Traditional two-stage SPR methods rely heavily on intermediate ASR transcription steps. To verify the performance of two-stage methods under different ASR models, we evaluate the baseline TBPR methods using transcripts by 4 different ASR models.
As shown in Tab.~\ref{tab:asr_comparison}, despite the use of different advanced ASR systems, all evaluated models still exhibit WERs exceeding 15\%, causing the retrieval accuracy of two-stage pipelines to remain consistently sub-optimal. 
In contrast, Pro-SPA directly aligns raw speech with visual appearance without relying on any intermediate transcription step. 
By bypassing this stage, our approach eliminates ASR recognition errors that distort or omit fine-grained appearance attributes, achieving state-of-the-art performance and outperforming all these evaluated ASR-based combinations. Inference efficiency comparisons are detailed in Appendix~\ref{app:inference_efficiency}.

\begin{table}[t]
    \centering
    \footnotesize
    \setlength{\abovecaptionskip}{3pt}
    \setlength{\belowcaptionskip}{3pt}
    \setlength{\tabcolsep}{4pt}
    \caption{\textbf{Comparison with two-stage ASR+TBPR baselines across different ASR models on Speech-PEDES.} WER denotes the Word Error Rate of each ASR system. Best results are highlighted in \textbf{bold}.}
    \label{tab:asr_comparison}
    \begin{tabular}{c|l||cccc}
        \hline\thickhline
        \rowcolor{headgray}
        ASR Model & Method & R-1 & R-5 & R-10 & mAP \\
        \hline\hline
        \multirow{9}{*}{\shortstack{wav2vec 2.0~\cite{wav2vec} \\ (WER: 15.98\%)}}
           & IRRA~\cite{Jiang2023IRRA}            & 54.53 & 73.62 & 81.06 & 50.29 \\
           & TBPS-CLIP~\cite{Cao2024TBPSCLIP}     & 55.05 & 73.36 & 80.34 & 49.89 \\
           & RDE~\cite{Qin2024RDE}                & 60.69 & 79.13 & 85.48 & 54.95 \\
           & DM-Adapter~\cite{Liu2025DMAdapter}   & 54.92 & 74.04 & 81.16 & 49.97 \\
           & HCA~\cite{hca}                       & 57.81 & 77.10 & 83.40 & 53.29 \\
           & GA-DMS~\cite{Zheng2025GADMS}         & 59.93 & 77.21 & 83.33 & 53.98 \\
           & MRA~\cite{MRA}                       & 60.61 & 77.86 & 84.23 & 54.41 \\
           & CECA~\cite{Zou2026CECA}              & 56.84 & 76.01 & 82.67 & 51.21 \\
           & OMRE~\cite{OMRE}                     & 56.76 & 75.54 & 82.72 & 51.20 \\
        \hline
        
        \multirow{9}{*}{\shortstack{HuBERT~\cite{hubert} \\ (WER: 17.25\%)}}
           & IRRA~\cite{Jiang2023IRRA}            & 52.99 & 72.66 & 80.47 & 49.33 \\
           & TBPS-CLIP~\cite{Cao2024TBPSCLIP}     & 53.78 & 72.37 & 78.64 & 48.98 \\
           & RDE~\cite{Qin2024RDE}                & 58.90 & 77.32 & 83.22 & 53.63 \\
           & DM-Adapter~\cite{Liu2025DMAdapter}   & 53.09 & 72.77 & 80.02 & 48.77 \\
           & HCA~\cite{hca}                       & 55.59 & 74.97 & 81.90 & 51.92 \\
           & GA-DMS~\cite{Zheng2025GADMS}         & 57.76 & 75.24 & 81.68 & 52.54 \\
           & MRA~\cite{MRA}                       & 58.61 & 76.75 & 83.15 & 53.15 \\
           & CECA~\cite{Zou2026CECA}              & 55.65 & 74.38 & 81.25 & 50.09 \\
           & OMRE~\cite{OMRE}                     & 55.13 & 74.17 & 81.16 & 50.13 \\
        \hline
        \multirow{9}{*}{\shortstack{Moonshine~\cite{Moonshine} \\ (WER: 18.72\%)}}
           & IRRA~\cite{Jiang2023IRRA}            & 53.23 & 72.30 & 79.30 & 49.44 \\
           & TBPS-CLIP~\cite{Cao2024TBPSCLIP}     & 53.33 & 71.12 & 77.53 & 48.42 \\
           & RDE~\cite{Qin2024RDE}                & 58.04 & 76.27 & 82.21 & 52.96 \\
           & DM-Adapter~\cite{Liu2025DMAdapter}   & 53.49 & 72.21 & 79.37 & 49.00 \\
           & HCA~\cite{hca}                       & 56.43 & 74.89 & 81.48 & 52.15 \\
           & GA-DMS~\cite{Zheng2025GADMS}         & 57.65 & 75.08 & 81.63 & 52.48 \\
           & MRA~\cite{MRA}                       & 57.88 & 75.36 & 81.84 & 52.49 \\
           & CECA~\cite{Zou2026CECA}              & 55.62 & 73.44 & 80.08 & 50.09 \\
           & OMRE~\cite{OMRE}                     & 55.17 & 74.27 & 81.03 & 50.13 \\
        \hline
        \multirow{9}{*}{\shortstack{Whisper~\cite{whisper} \\ (WER: 15.36\%)}}
           & IRRA~\cite{Jiang2023IRRA}            & 56.68 & 76.48 & 83.09 & 52.18 \\
           & TBPS-CLIP~\cite{Cao2024TBPSCLIP}     & 57.26 & 75.60 & 81.94 & 51.75 \\
           & RDE~\cite{Qin2024RDE}                & 61.61 & 79.99 & 85.66 & 55.07 \\
           & DM-Adapter~\cite{Liu2025DMAdapter}   & 57.00 & 75.96 & 83.01 & 51.84 \\
           & HCA~\cite{hca}                       & 59.67 & 78.15 & 84.37 & 54.83 \\
           & GA-DMS~\cite{Zheng2025GADMS}         & 60.92 & 79.27 & 84.86 & 55.02 \\
           & MRA~\cite{MRA}                       & 61.24 & 79.00 & 85.32 & 55.13 \\
           & CECA~\cite{Zou2026CECA}              & 57.98 & 76.59 & 83.27 & 52.11 \\
           & OMRE~\cite{OMRE}                     & 58.63 & 77.76 & 84.10 & 52.69 \\
        \hline
        \rowcolor[HTML]{D7F6FF}
        \multicolumn{2}{c||}{Pro-SPA (Ours)} & \textbf{63.42} & \textbf{81.90} & \textbf{88.37} & \textbf{55.21} \\
        \hline\thickhline
    \end{tabular}
    \vspace{-8pt}
\end{table}

\begin{figure}[t]
    \centering
    \setlength{\abovecaptionskip}{3pt}
    \setlength{\belowcaptionskip}{3pt}
    \includegraphics[width=\linewidth]{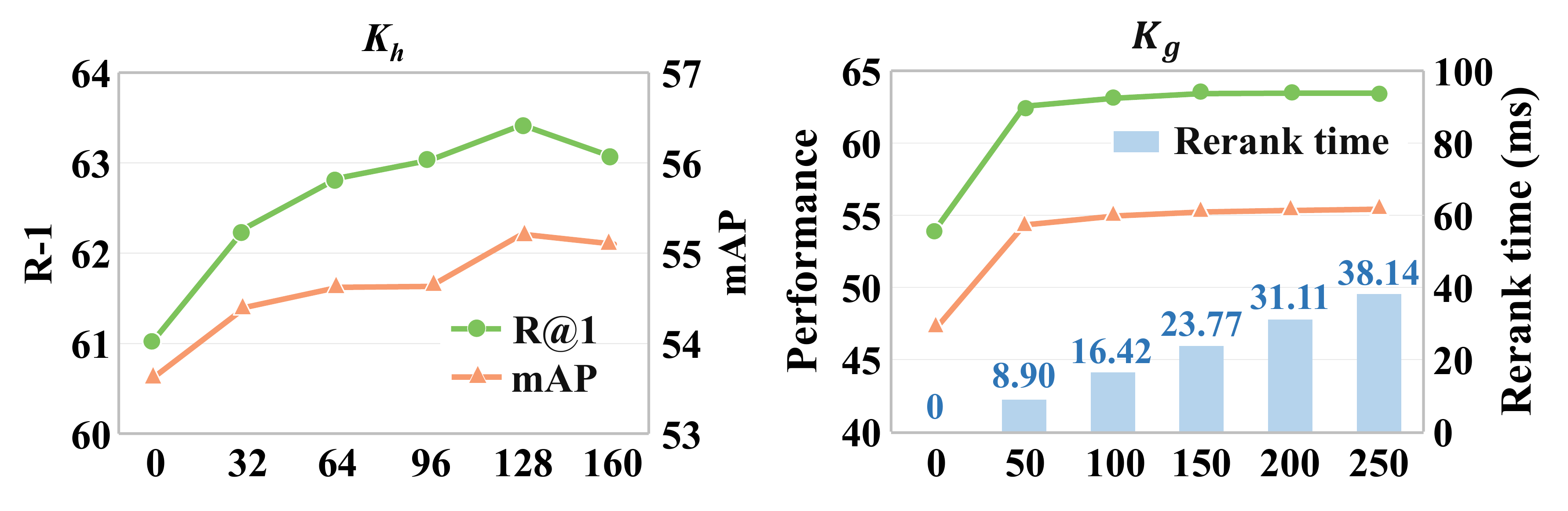}
    \caption{Ablation of hyperparameters $K_h$ and $K_g$. (Left) Performance under varying numbers of hard negatives $K_h$ for training LSSA. (Right) Performance under varying candidate set sizes $K_g$ for dual-granularity score calibration in DGSC.}
    \label{fig:hyper-parameter}
    \vspace{-10pt}
\end{figure}

\begin{figure}[t]
    \centering
    \setlength{\abovecaptionskip}{3pt}
    \setlength{\belowcaptionskip}{3pt}
    \includegraphics[width=.65\linewidth]{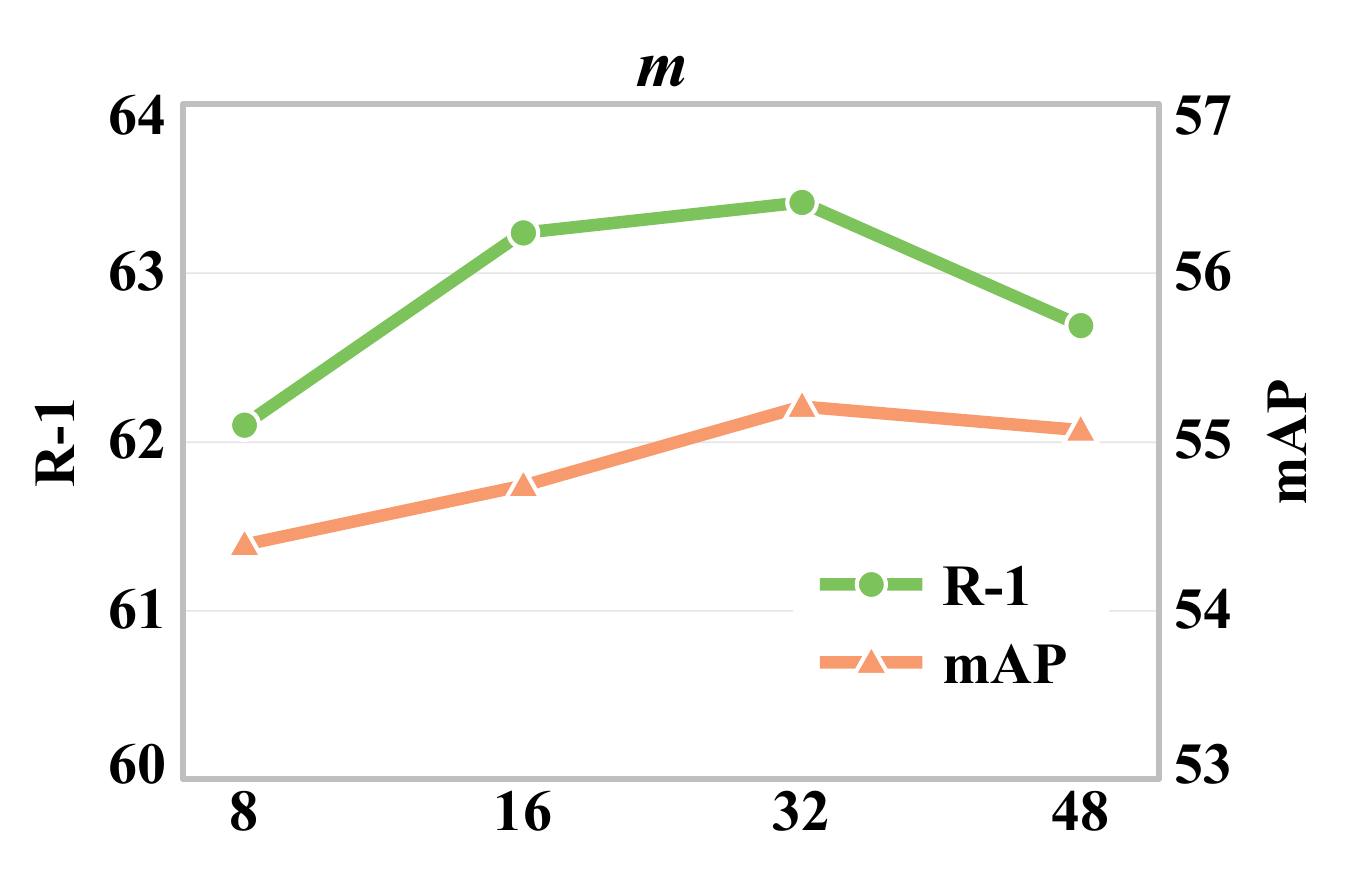}
    \caption{Effect of the number of speech segments~$m$ in LSSA.}
    \label{fig:additional-hyper-parameter}
    \vspace{-10pt}
\end{figure}

\subsection{Comparison of Contrastive Objectives}
\label{app:contrastive_objectives}

To evaluate the effectiveness of our Identity-aware Cross-modal Contrastive (ICC) objective, we compare it against two standard contrastive losses, SigLIP~\cite{siglip} and InfoNCE~\cite{infonce}, under the identical Pro-SPA backbone and training protocol. 
As reported in Tab.~\ref{tab:loss_comparison}, ICC consistently outperforms both baselines across all retrieval metrics. 
Compared with single-positive InfoNCE, ICC improves both R-1 and mAP, supporting the effectiveness of identity-aware supervision for speech-based person retrieval.
Corresponding qualitative visualizations are provided in Appendix~\ref{app:visualization_info_vs_ours}.

\begin{figure*}[t]
    \centering
    \setlength{\abovecaptionskip}{3pt}
    \setlength{\belowcaptionskip}{3pt}
    \includegraphics[width=\linewidth]{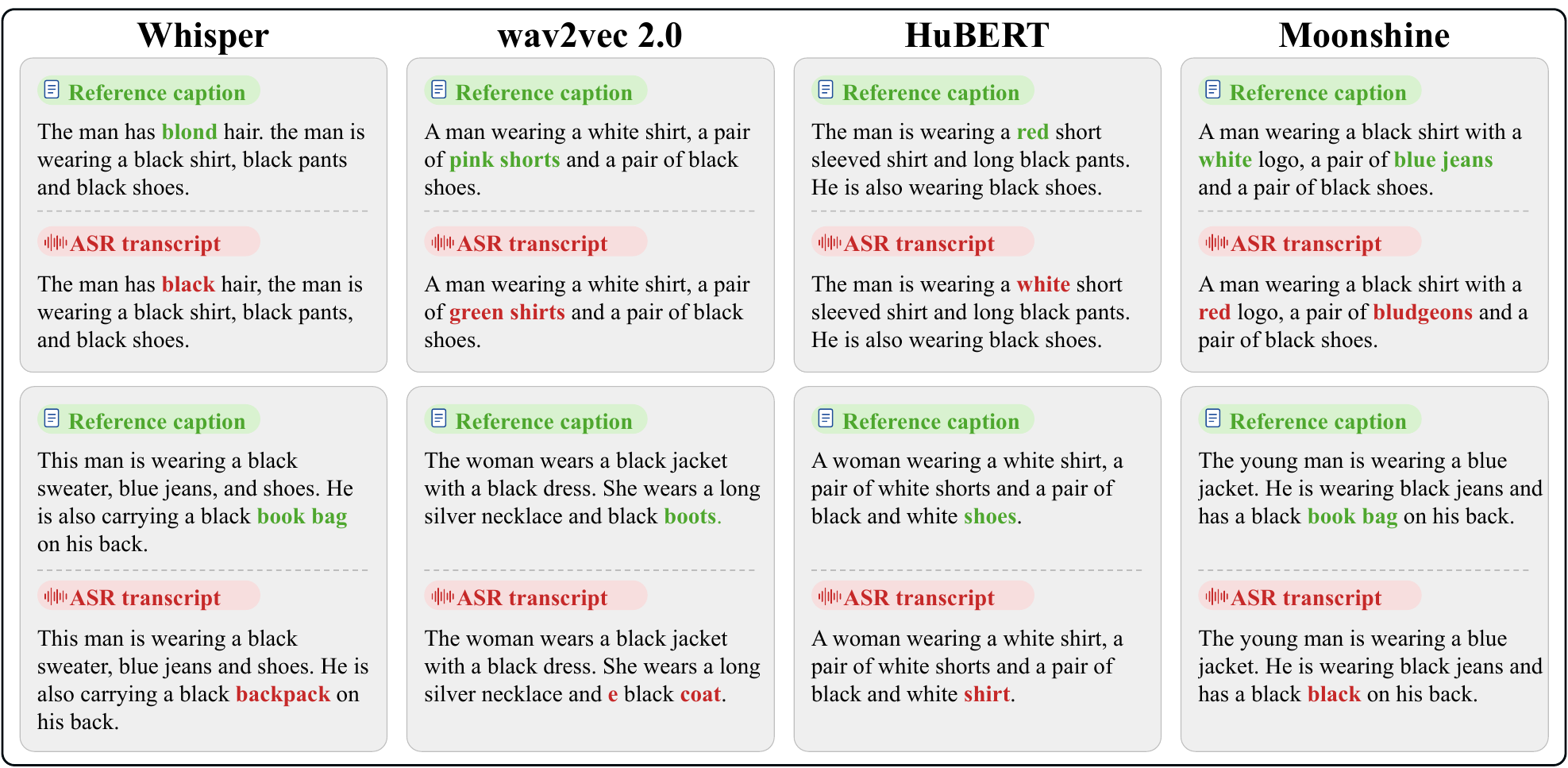}
    \caption{\textbf{Representative ASR transcription errors on Speech-PEDES.} Comparison between reference captions and transcripts generated by Whisper, wav2vec 2.0, HuBERT, and Moonshine. Erroneous or omitted attribute phrases are highlighted in red.}
    \label{fig:asr_errors}
    \vspace{-10pt}
\end{figure*}

\subsection{Hyper-Parameter Analysis}
\label{app:speech_segments}

Pro-SPA involves three key hyper-parameters: the number of hard negatives $K_h$ during LSSA training, the candidate set size $K_g$ for DGSC reranking, and the pooling parameter $m$. 
As shown in Fig.~\ref{fig:hyper-parameter}~(Left), performance peaks at $K_h = 128$ and then degrades slightly. This indicates that while hard negatives are useful, an excessive number may introduce noisy gradients that disturb fine-grained alignment.
Fig.~\ref{fig:hyper-parameter}~(Right) shows a performance gain as $K_g$ increases. Since the computational cost of local cross-attention also increases with $K_g$, we set $K_g = 150$ to strike a balance between accuracy and efficiency.
Fig.~\ref{fig:additional-hyper-parameter} shows that performance improves as $m$ increases from 8 to 32, but declines at $m=48$. This trend suggests that overly coarse pooling may merge distinct appearance cues, while overly fine segmentation may fragment them. We therefore use $m=32$ as the default setting.

\begin{table}[t]
    \centering
    \footnotesize
    \setlength{\abovecaptionskip}{3pt}
    \setlength{\belowcaptionskip}{3pt}
    \setlength{\tabcolsep}{5pt}
    \caption{\textbf{Comparison of contrastive learning objectives on Speech-PEDES.} All models share the same Pro-SPA architecture and training setup. Best results are in \textbf{bold}.}
    \label{tab:loss_comparison}
    \resizebox{\linewidth}{!}{%
    \begin{tabular}{l|c||cccc}
        \hline\thickhline
        \rowcolor{headgray}
        Objective & Alignment & R-1~$\uparrow$ & R-5~$\uparrow$ & R-10~$\uparrow$ & mAP~$\uparrow$ \\
        \hline\hline
        SigLIP~\cite{siglip} & Pairwise & 60.82 & 80.60 & 86.79 & 52.05 \\
        InfoNCE~\cite{infonce} & Instance-level & 61.71 & 81.64 & 88.34 & 53.43 \\
        \hline
        \rowcolor[HTML]{D7F6FF}
        ICC (Ours) & Identity-level & \textbf{63.42} & \textbf{81.90} & \textbf{88.37} & \textbf{55.21} \\
        \hline\thickhline
    \end{tabular}}
    \vspace{-8pt}
\end{table}

\begin{table}[t]
    \centering
    \footnotesize
    \setlength{\abovecaptionskip}{3pt}
    \setlength{\belowcaptionskip}{3pt}
    \setlength{\tabcolsep}{6pt}
    \caption{\textbf{Comparison of inference efficiency on the Speech-PEDES.} Total Time measures full test-set transcription (for ASR models) or end-to-end retrieval inference (for Pro-SPA), with relative latency normalized to Pro-SPA. The fastest result is in \textbf{bold}.}
    \label{tab:supp_efficiency}
    \resizebox{\linewidth}{!}{%
    \begin{tabular}{l|c||cc}
        \hline\thickhline
        \rowcolor{headgray}
        Model & Scope & Total Time (s)~$\downarrow$ & Rel. Time~$\downarrow$ \\
        \hline\hline
        wav2vec 2.0~\cite{wav2vec} & ASR Only & 256.6 & 1.69$\times$ \\
        HuBERT~\cite{hubert} & ASR Only & 245.5 & 1.62$\times$ \\
        Moonshine~\cite{Moonshine} & ASR Only & 1780.7 & 11.74$\times$ \\
        Whisper~\cite{whisper} & ASR Only & 559.7 & 3.69$\times$ \\
        \hline
        \rowcolor[HTML]{D7F6FF}
        Pro-SPA (Ours) & End-to-End & \textbf{151.7} & \textbf{1.00$\times$} \\
        \hline\thickhline
    \end{tabular}}
    \vspace{-8pt}
\end{table}

\subsection{Inference Efficiency}
\label{app:inference_efficiency}
Beyond the transcription errors analyzed in Appendix~\ref{app:asr_models}, intermediate ASR steps in two-stage pipelines also introduce interaction latency. To quantify this, we measure the total processing time of four representative ASR models on the Speech-PEDES test set and compare it against the inference time of our Pro-SPA, with all evaluations conducted on the same single NVIDIA GPU.

As shown in Tab.~\ref{tab:supp_efficiency}, the ASR transcription stage alone introduces substantial latency, even before the subsequent TBPR retrieval phase. 
In contrast, on a single NVIDIA A40 GPU, Pro-SPA completes retrieval over all 6,156 test queries in 151.7 s with a query batch size of 1, without intermediate transcription.
Under the evaluated configuration, Pro-SPA requires less total inference time than the transcription stage alone of the compared ASR systems.

\begin{table*}[t]
    \centering
    \setlength{\abovecaptionskip}{3pt}
    \setlength{\belowcaptionskip}{3pt}
    \caption{\textbf{Robustness comparison under Gaussian white noise on Speech-PEDES.} Percentages (0\%--30\%) denote the proportion of perturbed test queries (SNR $\in [0, 20]$\,dB), and $\Delta$R-1 indicates the R-1 change relative to the clean (0\%) setting. Best results (including the smallest performance drop) are in \textbf{bold}, and our Pro-SPA is highlighted in blue.}
    \small
    \resizebox{\linewidth}{!}{
        \renewcommand\arraystretch{1.2}
     \begin{tabular}{l|c||cccc|cccc:c|cccc:c|cccc:c}
    \hline\thickhline
        \rowcolor{headgray}
        & & \multicolumn{4}{c|}{0\% (WER: 15.36\%)} & \multicolumn{5}{c|}{10\% (WER: 18.03\%)} & \multicolumn{5}{c|}{20\% (WER: 20.09\%)} & \multicolumn{5}{c}{30\% (WER: 22.05\%)} \\
        \cline{3-21}
        \rowcolor{headgray}
        \multirow{-2}{*}{Methods} & \multirow{-2}{*}{Ref.} & R-1 & R-5 & R-10 & mAP & R-1 & R-5 & R-10 & mAP & $\Delta$R-1 & R-1 & R-5 & R-10 & mAP & $\Delta$R-1 & R-1 & R-5 & R-10 & mAP & $\Delta$R-1 \\
        \hline\hline
        IRRA~\cite{Jiang2023IRRA} & CVPR'23 
        & 56.68 & 76.48 & 83.09 & 52.18 
        & 54.73 & 74.14 & 80.77 & 50.55 & $-$1.95 
        & 52.79 & 71.75 & 78.44 & 48.87 & $-$3.89 
        & 50.94 & 69.66 & 76.33 & 47.27 & $-$5.74 \\
        
        TBPS-CLIP~\cite{Cao2024TBPSCLIP} & AAAI'24 
        & 57.26 & 75.60 & 81.94 & 51.75 
        & 55.26 & 73.16 & 79.65 & 50.06 & $-$2.00 
        & 53.25 & 70.92 & 77.37 & 48.33 & $-$4.01 
        & 51.40 & 68.76 & 75.16 & 46.74 & $-$5.86 \\
        
        RDE~\cite{Qin2024RDE} & CVPR'24 
        & 61.61 & 79.99 & 85.66 & 55.07 
        & 59.21 & 76.67 & 82.31 & 53.52 & $-$2.40 
        & 56.76 & 73.78 & 79.45 & 51.44 & $-$4.85 
        & 54.69 & 71.64 & 77.32 & 49.78 & $-$6.92 \\
        
        DM-Adapter~\cite{Liu2025DMAdapter} & AAAI'25 
        & 57.00 & 75.96 & 83.01 & 51.84 
        & 54.92 & 73.59 & 80.73 & 50.09 & $-$2.08 
        & 52.92 & 71.28 & 78.27 & 48.33 & $-$4.08 
        & 51.01 & 69.15 & 76.25 & 46.66 & $-$5.99 \\
        
        HCA~\cite{hca} & MM'25 
        & 59.67 & 78.15 & 84.37 & 54.83 
        & 57.55 & 75.78 & 81.99 & 53.05 & $-$2.12 
        & 55.43 & 73.39 & 79.58 & 51.25 & $-$4.24 
        & 53.46 & 71.02 & 77.34 & 49.50 & $-$6.21 \\
        
        GA-DMS~\cite{Zheng2025GADMS} & EMNLP'25 
        & 60.92 & 79.27 & 84.86 & 55.02 
        & 58.89 & 76.87 & 82.47 & 53.51 & $-$2.03 
        & 56.60 & 74.35 & 80.10 & 51.65 & $-$4.32 
        & 54.63 & 72.14 & 77.94 & 49.94 & $-$6.29 \\
        
        MRA~\cite{MRA} & PR'26
        & 61.24 & 79.00 & 85.32 & 55.13 
        & 59.23 & 76.59 & 82.88 & 53.53 & $-$2.01 
        & 57.05 & 74.20 & 80.47 & 51.72 & $-$4.19 
        & 55.05 & 72.11 & 78.41 & 50.03 & $-$6.19 \\
        
        CECA~\cite{Zou2026CECA} & CVPR'26 
        & 57.98 & 76.59 & 83.27 & 52.11 
        & 55.93 & 74.33 & 81.03 & 50.46 & $-$2.05 
        & 53.96 & 72.08 & 78.80 & 48.76 & $-$4.02 
        & 51.97 & 70.01 & 76.80 & 47.19 & $-$6.01 \\
        
        OMRE~\cite{OMRE} & TIFS'26 
        & 58.63 & 77.76 & 84.10 & 52.69 
        & 56.58 & 75.42 & 81.73 & 51.05 & $-$2.05 
        & 54.53 & 73.15 & 79.42 & 49.36 & $-$4.10 
        & 52.58 & 70.99 & 77.32 & 47.76 & $-$6.05 \\
        
        \hline
        \rowcolor[HTML]{D7F6FF}
        Pro-SPA (Ours) & -- 
        & \textbf{63.42} & \textbf{81.90} & \textbf{88.37} & \textbf{55.21} 
        & \textbf{62.93} & \textbf{81.58} & \textbf{88.04} & \textbf{54.86} & \textbf{$-$0.49} 
        & \textbf{62.39} & \textbf{81.12} & \textbf{87.67} & \textbf{54.42} & \textbf{$-$1.03} 
        & \textbf{61.97} & \textbf{80.70} & \textbf{87.25} & \textbf{54.06} & \textbf{$-$1.45} \\
    \hline\thickhline
    \end{tabular}
    }
    \label{tab:noise_gussian}
    \vspace{-10pt}
\end{table*}

\begin{table*}[t]
    \centering
    \setlength{\abovecaptionskip}{3pt}
    \setlength{\belowcaptionskip}{3pt}
    \caption{\textbf{Robustness comparison under environmental noise on Speech-PEDES.} Percentages (0\%--30\%) denote the proportion of perturbed test queries (SNR $\in [0, 20]$\,dB), and $\Delta$R-1 indicates the R-1 change relative to the clean (0\%) setting. Best results (including the smallest performance drop) are in \textbf{bold}, and our Pro-SPA is highlighted in blue.}
    \small
    \resizebox{\linewidth}{!}{
        \renewcommand\arraystretch{1.2}
    \begin{tabular}{l|c||cccc|cccc:c|cccc:c|cccc:c}
    \hline\thickhline
        \rowcolor{headgray}
        & & \multicolumn{4}{c|}{0\% (WER: 15.36\%)} & \multicolumn{5}{c|}{10\% (WER: 16.52\%)} & \multicolumn{5}{c|}{20\% (WER: 17.05\%)} & \multicolumn{5}{c}{30\% (WER: 17.59\%)} \\
        \cline{3-21}
        \rowcolor{headgray}
        \multirow{-2}{*}{Methods} & \multirow{-2}{*}{Ref.} & R-1 & R-5 & R-10 & mAP & R-1 & R-5 & R-10 & mAP & $\Delta$R-1 & R-1 & R-5 & R-10 & mAP & $\Delta$R-1 & R-1 & R-5 & R-10 & mAP & $\Delta$R-1 \\
        \hline\hline
        IRRA~\cite{Jiang2023IRRA} & CVPR'23 
        & 56.68 & 76.48 & 83.09 & 52.18 
        & 56.14 & 75.91 & 82.50 & 51.74 & $-$0.54 
        & 55.65 & 75.37 & 82.03 & 51.31 & $-$1.03 
        & 55.21 & 74.81 & 81.30 & 50.90 & $-$1.47 \\
        
        TBPS-CLIP~\cite{Cao2024TBPSCLIP} & AAAI'24 
        & 57.26 & 75.60 & 81.94 & 51.75 
        & 56.82 & 75.03 & 81.58 & 51.30 & $-$0.44 
        & 56.19 & 74.42 & 80.98 & 50.75 & $-$1.07 
        & 55.78 & 73.93 & 80.49 & 50.35 & $-$1.48 \\
        
        RDE~\cite{Qin2024RDE} & CVPR'24 
        & 61.61 & 79.99 & 85.66 & 55.07 
        & 61.13 & 79.17 & 84.93 & 54.84 & $-$0.48 
        & 60.67 & 78.59 & 84.50 & 54.77 & $-$0.94 
        & 60.19 & 77.89 & 83.79 & 54.33 & $-$1.42 \\
        
        DM-Adapter~\cite{Liu2025DMAdapter} & AAAI'25 
        & 57.00 & 75.96 & 83.01 & 51.84 
        & 56.50 & 75.32 & 82.49 & 51.41 & $-$0.50 
        & 56.06 & 74.76 & 81.89 & 50.93 & $-$0.94 
        & 55.69 & 74.16 & 81.42 & 50.53 & $-$1.31 \\
        
        HCA~\cite{hca} & MM'25 
        & 59.67 & 78.15 & 84.37 & 54.83 
        & 59.11 & 77.73 & 83.87 & 54.42 & $-$0.56 
        & 58.33 & 77.18 & 83.38 & 53.85  & $-$1.34 
        & 57.98 & 76.59 & 82.76 & 53.48 & $-$1.69 \\
        
        GA-DMS~\cite{Zheng2025GADMS} & EMNLP'25 
        & 60.92 & 79.27 & 84.86 & 55.02 
        & 60.43 & 78.69 & 84.28 & 54.78 & $-$0.49 
        & 59.70 & 77.92 & 83.66 & 54.24 & $-$1.22 
        & 59.42 & 77.42 & 83.11 & 53.91 & $-$1.50 \\
        
        MRA~\cite{MRA} & PR'26
        & 61.24 & 79.00 & 85.32 & 55.13 
        & 60.62 & 78.27 & 84.58 & 54.80 & $-$0.62 
        & 60.09 & 77.78 & 83.97 & 54.29 & $-$1.15 
        & 59.47 & 77.11 & 83.37 & 53.81 & $-$1.77 \\
        
        CECA~\cite{Zou2026CECA} & CVPR'26 
        & 57.98 & 76.59 & 83.27 & 52.11 
        & 57.62 & 76.19 & 82.85 & 51.70 & $-$0.36 
        & 57.20 & 75.68 & 82.44 & 51.25 & $-$0.78 
        & 56.73 & 75.06 & 82.00 & 50.87 & $-$1.25 \\
        
        OMRE~\cite{OMRE} & TIFS'26 
        & 58.63 & 77.76 & 84.10 & 52.69 
        & 58.12 & 77.24 & 83.63 & 52.29 & $-$0.51 
        & 57.63 & 76.82 & 83.20 & 51.77 & $-$1.00 
        & 57.10 & 76.40 & 82.80 & 51.40 & $-$1.53 \\
        
        \hline
        \rowcolor[HTML]{D7F6FF}
        Pro-SPA (Ours) & -- 
        & \textbf{63.42} & \textbf{81.90} & \textbf{88.37} & \textbf{55.21} 
        & \textbf{63.26} & \textbf{81.94} & \textbf{88.14} & \textbf{55.10} & \textbf{$-$0.16} 
        & \textbf{63.08} & \textbf{81.82} & \textbf{88.04} & \textbf{54.95} & \textbf{$-$0.34} 
        & \textbf{62.78} & \textbf{81.68} & \textbf{88.08} & \textbf{54.84} & \textbf{$-$0.64} \\
    \hline\thickhline
    \end{tabular}
    }
    \label{tab:noise_env}
    \vspace{-10pt}
\end{table*}

\subsection{Robustness to Speech Variations}
\label{app:robutness_to_vary}


\subsubsection{ASR Transcription Errors}
\label{app:asr_noise}


Fig.~\ref{fig:asr_errors} illustrates representative transcription errors produced by four ASR models (Whisper~\cite{whisper}, wav2vec 2.0~\cite{wav2vec}, HuBERT~\cite{hubert}, and Moonshine~\cite{Moonshine}) on Speech-PEDES. 
These failures fall into two primary patterns that directly degrade downstream TBPR. 
First, ASR models frequently substitute fine-grained color or clothing attributes with grammatically fluent yet visually contradictory words, such as transcribing \textit{``blond hair''} as \textit{``black hair''} (Whisper) or \textit{``red shirt''} as \textit{``white shirt''} (HuBERT). Since these corrupted transcripts remain syntactically valid, downstream TBPR models blindly accept these errors and retrieve mismatched identities.
Second, acoustic ambiguity can trigger phonetic distortions and semantic omissions, such as corrupting \textit{``blue jeans''} into \textit{``bludgeons''}, as well as turning \textit{``black boots''} into \textit{``e black coat''} (wav2vec 2.0). 
These examples illustrate how ASR transcription errors can propagate to downstream retrieval and motivate direct speech-based person retrieval.

\subsubsection{Robustness Against Background Noise}
\label{app:noise_robustness}
To evaluate the robustness of our model in complex real-world scenarios, we introduce simulated background noise into the evaluation. We employ two distinct types of acoustic interference: synthetic Gaussian white noise and realistic environmental noise sampled from the DEMAND~\cite{demand} dataset. 
Specifically, we randomly select 10\%, 20\%, and 30\% of the speech queries in the test set and inject noise with a signal-to-noise ratio (SNR) uniformly sampled between 0 and 20\,dB. 
To ensure a strictly fair comparison, we re-transcribe the perturbed speech using the same ASR model to accurately capture the transcription errors induced by the background noise.

The results under Gaussian and environmental noise are presented in Tab.~\ref{tab:noise_gussian} and Tab.~\ref{tab:noise_env}, respectively. 
As observed, traditional two-stage methods experience severe performance degradation when exposed to noise. 
For instance, at a 30\% Gaussian noise ratio, the R-1 accuracy of the strongest baseline (RDE) drops drastically by 6.92\%. 
The increased WER and reduced retrieval accuracy indicate that noise-induced transcription errors can adversely affect the evaluated ASR+TBPR pipelines.
In contrast, even at a 30\% noisy query ratio, Pro-SPA maintains competitive performance, with modest R-1 drop of only 1.45\% (Gaussian) and 0.64\% (Environmental). 
Under the evaluated noise settings, Pro-SPA exhibits smaller overall performance drops than the compared ASR+TBPR pipelines, providing evidence of improved noise robustness in these experiments.


\subsubsection{Robustness to Speaker Variations}
\label{app:robust_to_speaker}
To evaluate the sensitivity of Pro-SPA to speaker variations, we conduct a speaker-replacement experiment while keeping the textual content fixed. 
Specifically, we randomly select 10\%, 20\%, and 30\% of the test queries and replace their speaker voices with unseen voices from the GLOBE~\cite{globe} dataset, while keeping all other evaluation settings unchanged.
As shown in Tab.~\ref{tab:speakers}, Pro-SPA maintains stable overall retrieval performance across the evaluated replacement ratios.
Even when 30\% of the test queries undergo speaker replacement, the overall performance decreases only slightly relative to the original setting.
These results provide evidence of robustness to speaker variations under the evaluated replacement protocol.

To further examine the contribution of identity-aware supervision to speaker robustness, we compare ICC objective with standard single-positive InfoNCE under the same speaker-replacement protocol in Appendix~\ref{app:icc_speakers}.



\subsubsection{Speaker Robustness of ICC vs.\ InfoNCE}
\label{app:icc_speakers}

To examine the speaker robustness of conventional single-positive contrastive learning, Tab.~\ref{tab:icc_speakers} compares InfoNCE with our Identity-aware Cross-modal Contrastive (ICC) objective under the 30\% speaker replacement setting. 

Specifically, we follow the same replacement protocol as in Appendix~\ref{app:robust_to_speaker}.
The results show that even on the original test set (0\%), InfoNCE already trails ICC by 1.71\% in R-1. 
More critically, once 30\% of the test queries undergo speaker replacement, InfoNCE suffers a noticeable R-1 degradation of 1.77\% (over $6\times$ the drop observed with ICC).
Overall, ICC maintains much stronger stability under speaker shifts, supporting the effectiveness of identity-aware supervision for speaker robustness.


\begin{table}[t]
    \centering
    \setlength{\abovecaptionskip}{3pt}
    \setlength{\belowcaptionskip}{3pt}
    \caption{\textbf{Robustness of Pro-SPA to speaker variations on Speech-PEDES.} Speakers are randomly replaced for 10\%--30\% of the test queries. The original (0\%) setting is highlighted in blue.}
    \footnotesize
    \resizebox{\linewidth}{!}{
        \renewcommand\arraystretch{1.2}      
    \begin{tabular}{c||cccc}
    \hline\thickhline
    \rowcolor{headgray}
    Change Ratio & R-1 & R-5 & R-10 & mAP \\
    \hline\hline
    \rowcolor[HTML]{D7F6FF}
    0\%(Original speakers)  & 63.42 & 81.90 & 88.37 & 55.21 \\
    \hline
    10\% & 63.35 & 81.79 & 88.27 & 55.17 \\
    20\% & 63.30 & 81.74 & 88.17 & 55.09 \\
    30\% & 63.14 & 81.73 & 88.11 & 54.98 \\
    \hline\thickhline
    \end{tabular}
    }
    \label{tab:speakers}
    \vspace{-10pt}
\end{table}

\begin{table}[t]
    \centering
    \setlength{\abovecaptionskip}{3pt}
    \setlength{\belowcaptionskip}{3pt}
    \caption{\textbf{Robustness comparison between ICC and InfoNCE under speaker variations on Speech-PEDES.} Ratio denotes the proportion of speaker-replaced test queries, and $\Delta$R-1 indicates the R-1 change from 0\% to 30\%. Best results under each ratio are in \textbf{bold}, and our ICC loss is highlighted in blue.}
    \label{tab:icc_speakers}
    \footnotesize
    \renewcommand{\arraystretch}{1.2}
    \resizebox{\linewidth}{!}{
        \begin{tabular}{l|c||cccc|c}
            \hline\thickhline
            \rowcolor{headgray}
            Loss & Ratio & R-1 & R-5 & R-10 & mAP & $\Delta$R-1 \\
            \hline\hline
            & 0\%  & 61.71 & 81.64 & 88.34 & 53.43 & \\
            \multirow{-2}{*}{InfoNCE~\cite{infonce}}
            & 30\% & 59.94 & 80.51 & 86.87 & 52.79 & \multirow{-2}{*}{$-$1.77} \\
            \hline
            \rowcolor[HTML]{D7F6FF}
            & 0\%  & \textbf{63.42} & \textbf{81.90} & \textbf{88.37} & \textbf{55.21} & \\
            \rowcolor[HTML]{D7F6FF}
            \multirow{-2}{*}{ICC (Ours)}
            & 30\% & \textbf{63.14} & \textbf{81.73} & \textbf{88.11} & \textbf{54.98} & \multirow{-2}{*}{\textbf{$-$0.28}} \\
            \hline\thickhline
        \end{tabular}
    }
    \vspace{-10pt}
\end{table}

\begin{table}[t]
    \centering
    \setlength{\abovecaptionskip}{3pt}
    \setlength{\belowcaptionskip}{3pt}
    \caption{\textbf{Effect of negative sampling strategies in LSSA.} We compare our hard negative mining (Pro-SPA) against using no additional negatives (Baseline) and randomly sampled negatives (Random). \textbf{Bold} indicates the best performance.}
    \footnotesize
    \resizebox{\linewidth}{!}{
        \renewcommand\arraystretch{1.2}
    \begin{tabular}{l|c||cccc}
    \hline\thickhline
        \rowcolor{headgray}
        Negative Strategy & $K_h$ & R-1 & R-5 & R-10 & mAP \\
        \hline\hline
        Baseline        & - & 61.03 & 80.57 & 87.26 & 53.62 \\
        Random          & 128 & 60.77 & 80.38 & 87.38 & 53.54 \\
        \hline
        \rowcolor[HTML]{D7F6FF}
        Pro-SPA & 128 & \textbf{63.42} & \textbf{81.90} & \textbf{88.37} & \textbf{55.21} \\
    \hline\thickhline
    \end{tabular}
    }
    \label{tab:negatives}
    \vspace{-8pt}
\end{table}

\begin{table}[t]
    \centering
    \setlength{\abovecaptionskip}{3pt}
    \setlength{\belowcaptionskip}{3pt}
    \caption{\textbf{Comparison of aggregation strategies in LSSA.} We compare our Bidirectional Consistency Aggregation (BCA) against standard aggregation strategies. \textbf{Bold} indicates the best performance, and our final configuration is highlighted in blue.}
    \footnotesize
    \resizebox{\linewidth}{!}{
        \renewcommand\arraystretch{1.2}
    \begin{tabular}{l||cccc}
    \hline\thickhline
        \rowcolor{headgray}
        Aggregation Strategy & R-1 & R-5 & R-10 & mAP \\
        \hline\hline
        Baseline (ISA-only)  & 54.01 & 76.04 & 83.74 & 47.13 \\
        Mean Pooling         & 60.56 & 80.20 & 86.66 & 53.15 \\
        Attention Pooling    & 61.78 & 81.35 & 87.88 & 54.02 \\
        Max Pooling          & 62.83 & 81.77 & 88.17 & 55.19 \\
        \hline
        \rowcolor[HTML]{D7F6FF}
        BCA (Ours)           & \textbf{63.42} & \textbf{81.90} & \textbf{88.37} & \textbf{55.21} \\
    \hline\thickhline
    \end{tabular}
    }
    \label{tab:aggregation}
    \vspace{-10pt}
\end{table}

\subsection{Effect of negative sampling strategies in LSSA}
\label{app:hn_effect_lssa}

To investigate how negative sampling impacts Local Speech-Segment Alignment (LSSA), we compare our hard negative mining strategy against two alternatives: using no additional negatives (Baseline) and sampling an equal number of random negatives (Random, $K_h = 128$).

The results in Tab.~\ref{tab:negatives} show that introducing 128 randomly sampled negatives does not consistently improve over the baseline: R-10 increases slightly, while R-1, R-5, and mAP decrease.
In contrast, our hard negative mining strategy yields greater improvements across all metrics. 
These results support the effectiveness of hard negative mining in LSSA over random sampling for improving cross-modal retrieval~performance.

\subsection{Effect of aggregation strategies in LSSA}
\label{app:compare_agg_lssa}

To validate the design of our Bidirectional Consistency Aggregation (BCA) in LSSA, we replace BCA with other standard pooling strategies (\emph{e.g.}, Mean Pooling, Attention Pooling, and Max Pooling), while keeping all other components and training settings~unchanged.

As shown in Tab.~\ref{tab:aggregation}, all local aggregation variants outperform the baseline, confirming the importance of fine-grained feature aggregation. 
Among them, Mean Pooling yields the lowest accuracy, as averaging over all units dilutes discriminative appearance cues with non-informative speech frames and visual background regions. 
While Attention Pooling and Max Pooling further improve performance, they still fall short of our proposed method. 
In contrast, by adaptively assigning greater weight to units with stronger reciprocal cross-modal support, BCA effectively highlights identity-discriminative cues and achieves the highest performance across all metrics, validating the effectiveness of our BCA design.

\begin{figure*}[t]
    \centering
    \setlength{\abovecaptionskip}{3pt}
    \setlength{\belowcaptionskip}{3pt}
    \includegraphics[width=\linewidth]{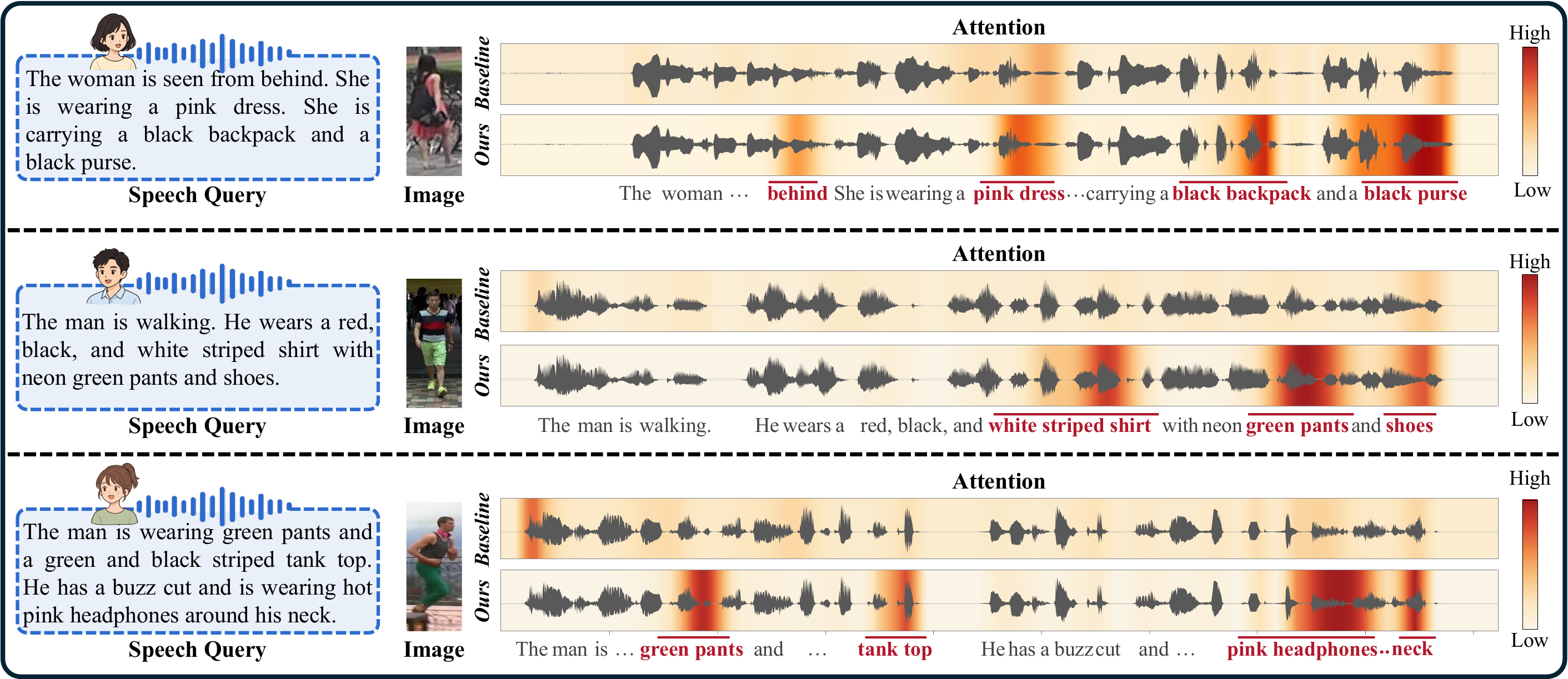}
    \caption{\textbf{Speech attention visualization of the baseline (WAVE) and Pro-SPA.} Background shading along the speech timeline indicates temporal attention weights (ranging from light to dark red, where darker shades denote stronger focus), with aligned words positioned at their timestamps and appearance-related phrases highlighted in red.}
    \label{fig:ours_vs_baseline}
    \vspace{-4pt}
\end{figure*}

\section{Visualizations}
\label{app:visualization}

\subsection{Attention Visualization}
\label{app:attention_visualization}

\subsubsection{Speech Attention: Pro-SPA vs.\ Baseline}
\label{app:local_alignment}

To examine how Local Speech-Segment Alignment (LSSA) captures fine-grained appearance cues without explicit word-level segmentation, we visualize the speech attention distributions of the baseline (WAVE~\cite{Tang2026WAVE}) and Pro-SPA across representative queries. 
As shown in Fig.~\ref{fig:ours_vs_baseline}, WAVE tends to distribute attention broadly and diffusely across the utterance, whereas Pro-SPA exhibits more concentrated attention on speech segments corresponding to discriminative pedestrian attributes.
Overall, these examples provide qualitative evidence that Pro-SPA focuses temporal attention on sparse appearance cues even in continuous, unsegmented speech, aligning with the intended design of LSSA.


\subsubsection{Cross-Modal Attention: ICC vs.\ InfoNCE}
\label{app:visualization_info_vs_ours}

\begin{figure*}[t]
    \centering
    \setlength{\abovecaptionskip}{3pt}
    \setlength{\belowcaptionskip}{3pt}
    \includegraphics[width=0.9\linewidth]{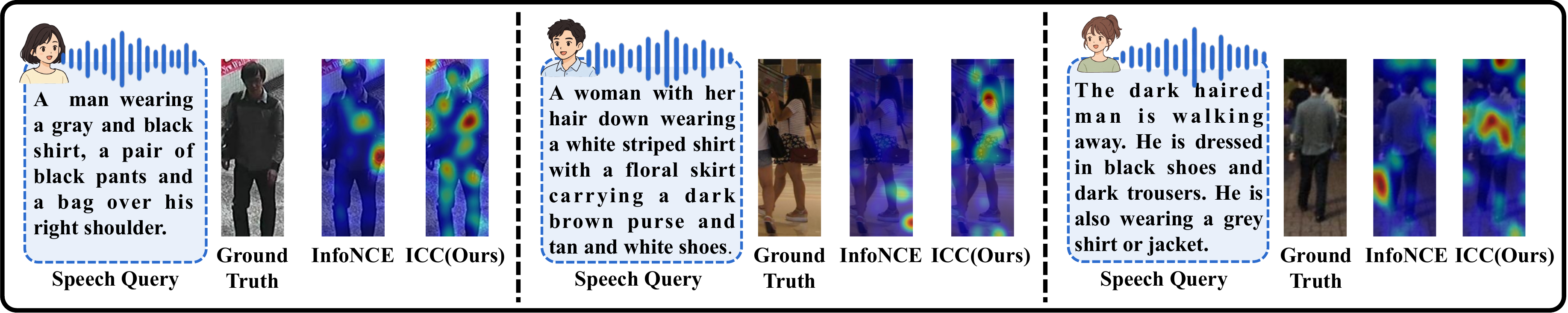}
    \caption{\textbf{Visualization of image attention under InfoNCE and ICC.}}
    \label{fig:infonce_vs_icc}
    \vspace{-10pt}
\end{figure*}

To qualitatively examine the effect of the contrastive objective, we compare the cross-modal attention  across both image and speech modalities for models trained with InfoNCE and our ICC objective on the same speech--image pairs.
As shown in Fig.~\ref{fig:infonce_vs_icc}, the ICC-trained model exhibits more concentrated visual attention on pedestrian appearance regions, whereas the InfoNCE-trained model also highlights background and attribute-irrelevant regions. 

\begin{figure*}[t]
    \centering
    \setlength{\abovecaptionskip}{3pt}
    \setlength{\belowcaptionskip}{3pt}
    \includegraphics[width=0.9\linewidth]{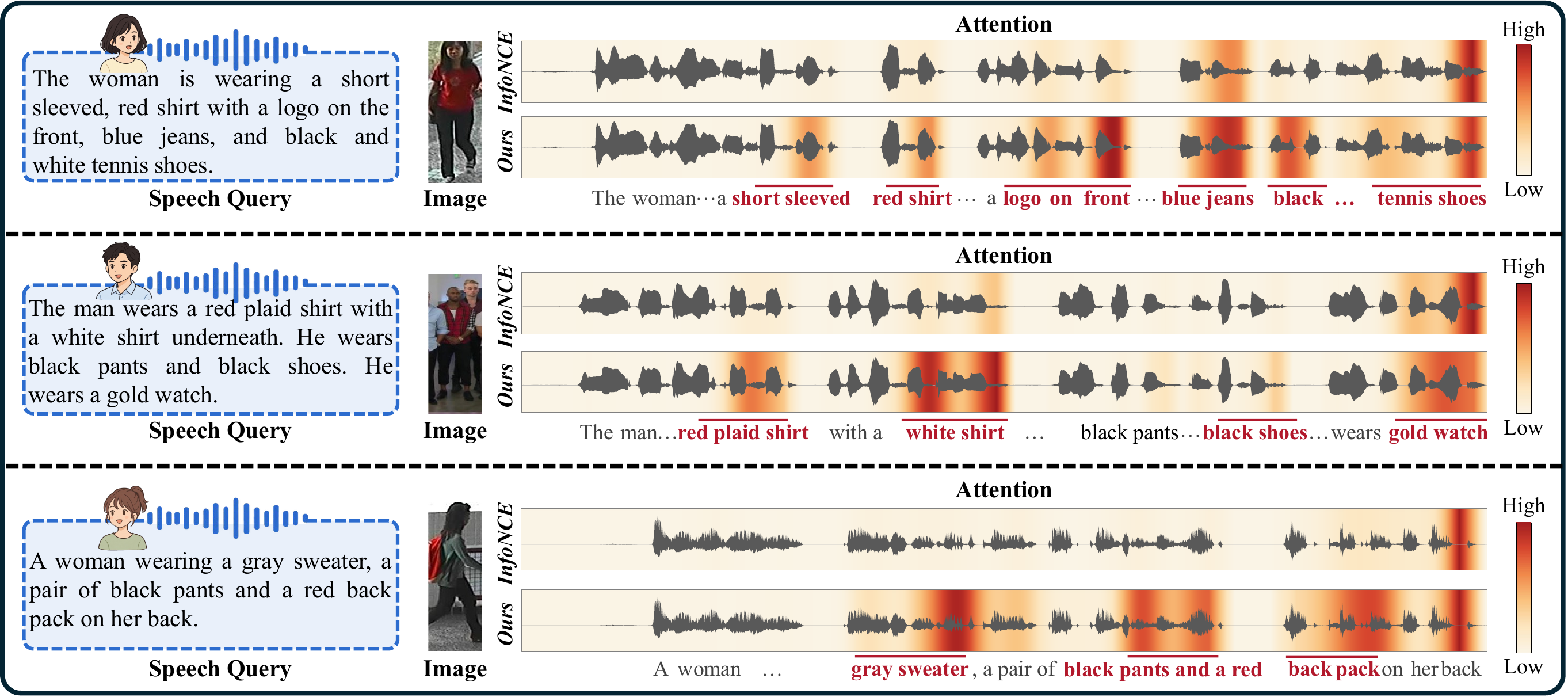}
    \caption{\textbf{Speech attention visualization under InfoNCE and ICC.} Background shading indicates attention weights (light to dark red), with aligned words positioned at their timestamps and attribute phrases highlighted in red.}
    \label{fig:attention-info}
    \vspace{-10pt}
\end{figure*}

Consistently, Fig.~\ref{fig:attention-info} shows that ICC produces more concentrated BCA weights on appearance-related speech segments while suppressing irrelevant words, whereas InfoNCE displays a more scattered temporal focus.
Together, these visualizations provide qualitative evidence that ICC anchors cross-modal alignment more tightly on discriminative appearance cues than single-positive InfoNCE; corresponding quantitative evidence of speaker robustness is reported in Tab.~\ref{tab:speakers} and Tab.~\ref{tab:icc_speakers}.

\subsection{Visualization of Retrieval Rankings}
\label{app:visualization_of_rank}

\begin{figure}[t]
    \centering
    \setlength{\abovecaptionskip}{3pt}
    \setlength{\belowcaptionskip}{3pt}
    \includegraphics[width=\linewidth]{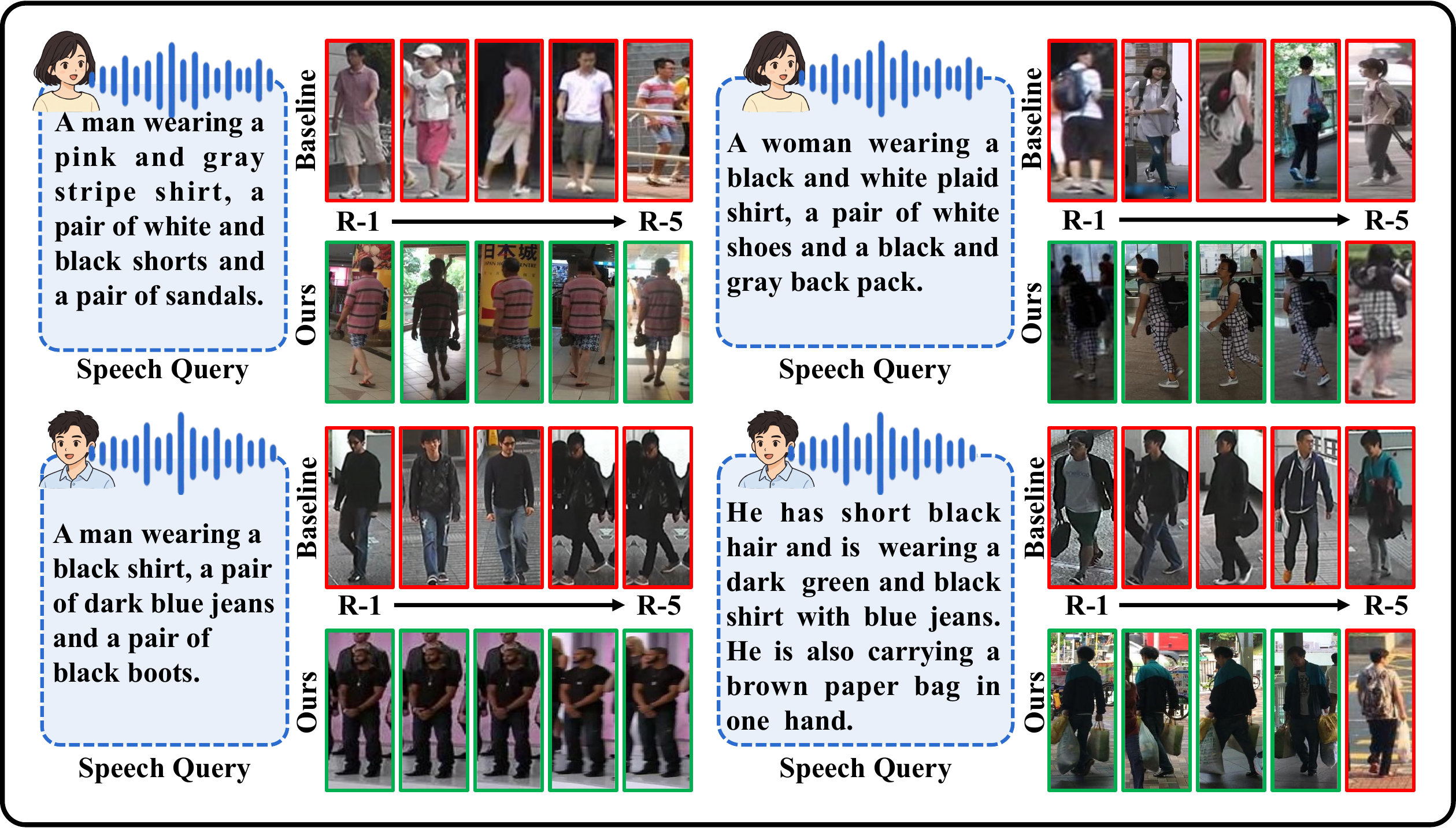}
    \caption{\textbf{Additional visualization of retrieval rankings.} Comparison between the baseline and Pro-SPA.}
    \label{fig:additional-rank}
    \vspace{-10pt}
\end{figure}

As a supplement to the main text, Fig.~\ref{fig:additional-rank} provides four additional retrieval ranking examples comparing Pro-SPA against the baseline.
Across these queries, Pro-SPA consistently ranks the target pedestrian higher, further demonstrating its superior retrieval performance.

\subsection{Visualization of Retrieval Process}
\label{app:visualization_retrieval}

\begin{figure*}[t]
    \centering
    \setlength{\abovecaptionskip}{3pt}
    \setlength{\belowcaptionskip}{3pt}
    \includegraphics[width=0.9\linewidth]{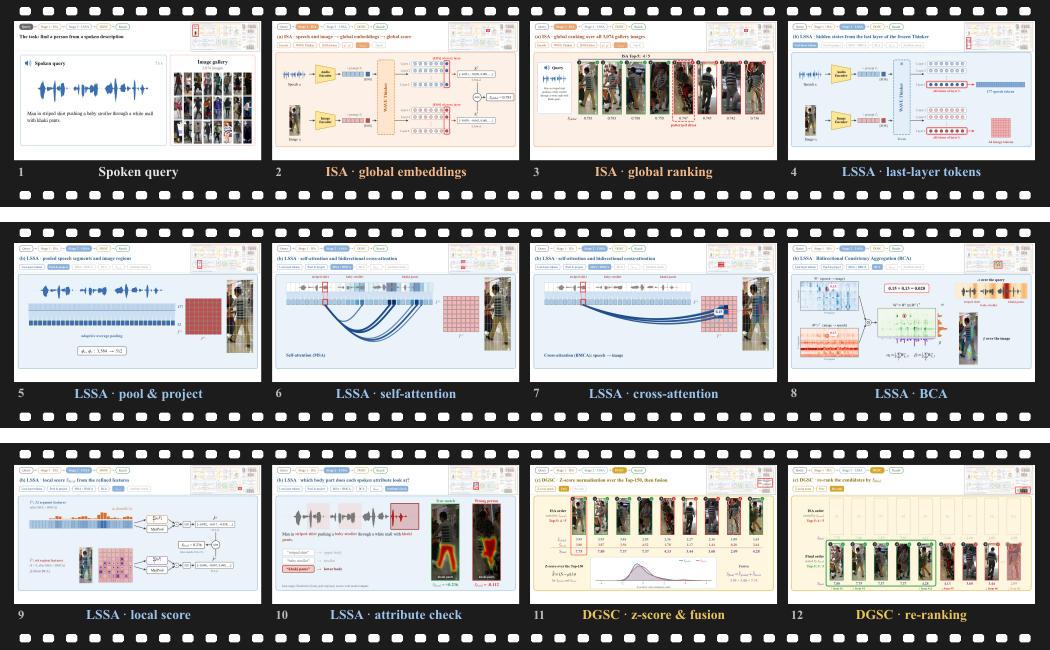}
    \caption{\textbf{Step-by-step key frames of the Pro-SPA retrieval process.} Captured from our narrated walkthrough video on a Speech-PEDES test query, illustrating the complete workflow from raw speech input and initial candidate retrieval (ISA) to fine-grained local re-scoring (LSSA) and final re-ranked results (DGSC).}
    \label{fig:filmstrip}
    \vspace{-12pt}
\end{figure*}

\begin{figure*}[t]
    \centering
    \includegraphics[width=\linewidth]{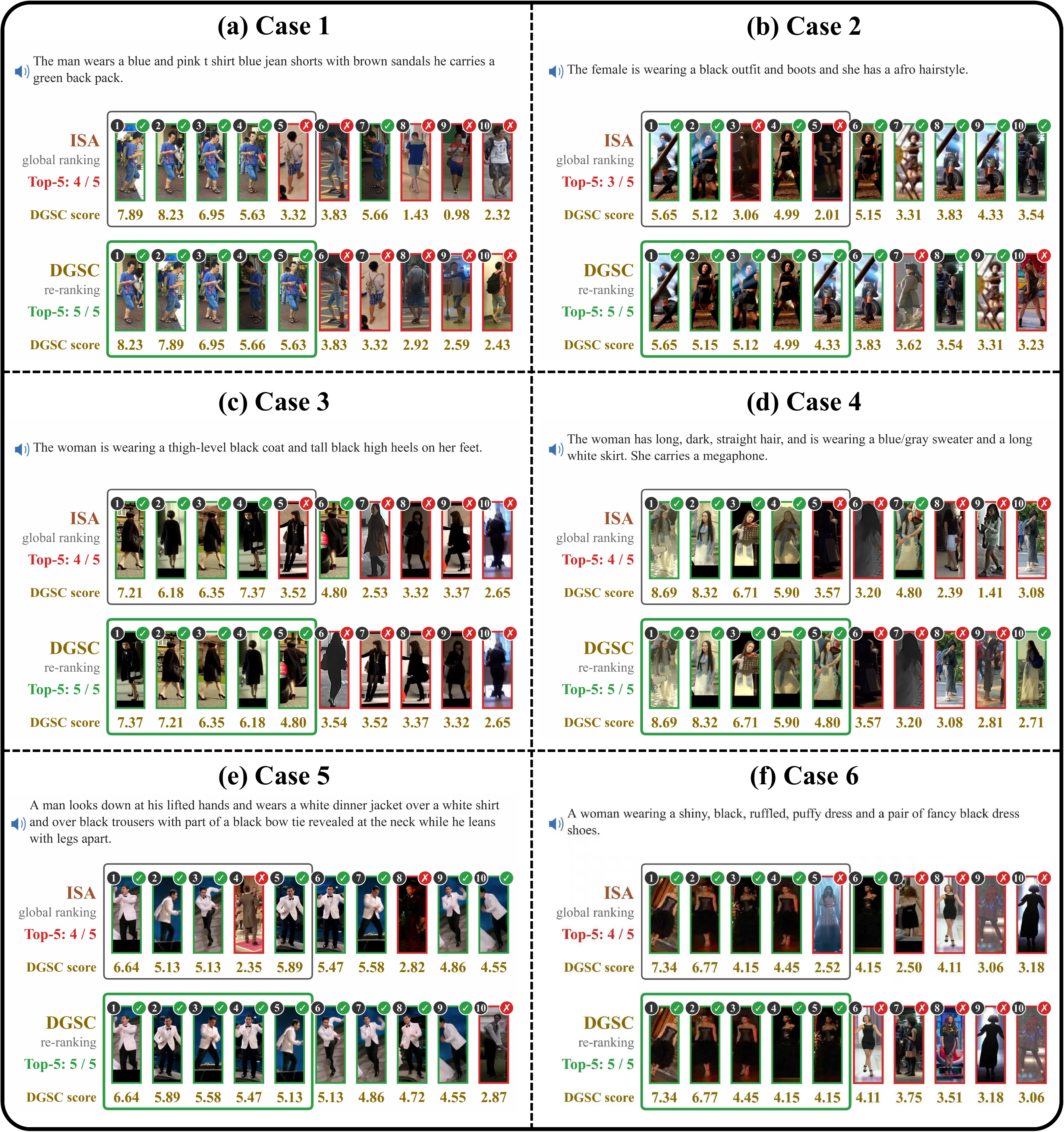}
    \caption{\textbf{Final retrieval result views across 6 Speech-PEDES test queries.} Each panel shows the concluding screen of a demo video, presenting the input speech query alongside the top-ranked pedestrian candidates retrieved by Pro-SPA.}
    \label{fig:retrieval-demo}
    \vspace{-10pt}
\end{figure*}

To illustrate the direct speech--image retrieval process, we provide six retrieval demo videos and one narrated walkthrough video in our \href{https://anonymous.4open.science/r/SPR-0365/}{anonymous repository}. 
Specifically, Figure~\ref{fig:retrieval-demo} shows the final result frames of the six demo videos, each corresponding to a speech query from the Speech-PEDES test set. 
In addition, Figure~\ref{fig:filmstrip} presents key frames from the narrated walkthrough video, tracing how Pro-SPA processes a single test query step by step to retrieve the target pedestrian.

\end{document}